\documentclass[a4paper,fleqn]{cas-dc}

\usepackage[numbers]{natbib}

\usepackage{xcolor}
\usepackage{tabularx}
\usepackage{booktabs}
\usepackage{array}

\usepackage{enumitem}
\setlist{nosep}

\usepackage{graphicx}
\usepackage{subcaption}

\usepackage{algorithm}
\usepackage{algorithmicx}
\usepackage{algpseudocode}

\begin{document}
\let\WriteBookmarks\relax
\def\floatpagepagefraction{1}
\def\textpagefraction{.001}

\newcommand{\revise}[1]{#1}

\makeatletter
\DeclareRobustCommand\onedot{\futurelet\@let@token\@onedot}
\def\@onedot{\ifx\@let@token.\else.\null\fi\xspace}

\def\eg{\emph{e.g}\onedot} \def\Eg{\emph{E.g}\onedot}
\def\ie{\emph{i.e}\onedot} \def\Ie{\emph{I.e}\onedot}
\def\cf{\emph{c.f}\onedot} \def\Cf{\emph{C.f}\onedot}
\def\etc{\emph{etc}\onedot} \def\vs{\emph{vs}\onedot}
\def\wrt{w.r.t\onedot} \def\dof{d.o.f\onedot}
\def\etal{\emph{et~al}\onedot}


\shorttitle{ShowFlow: From Robust Single Concept to Condition-Free Multi-Concept Generation}    
\shortauthors{Trong-Vu Hoang et al.}  

\title [mode = title]{ShowFlow: From Robust Single Concept to Condition-Free Multi-Concept Generation} 

\author[1,2]{Trong-Vu Hoang}[orcid=0000-0001-7367-1401] \ead{htvu@selab.hcmus.edu.vn}
\cormark[1]
\credit{Methodology, Conceptualization, Writing – original draft}

\author[1,2]{Quang-Binh Nguyen}[orcid=0000-0003-1199-3661] \ead{nqbinh@selab.hcmus.edu.vn}
\cormark[1]
\credit{Methodology, Conceptualization, Writing – original draft}

\author[3]{Thanh-Toan Do}[orcid=0000-0002-6249-0848] \ead{toan.do@monash.edu}
\credit{Methodology, Writing – Review \& Editing}

\author[4]{Tam V. Nguyen}[orcid=0000-0003-0236-7992] \ead{tamnguyen@udayton.edu}
\credit{Conceptualization, Writing – Review \& Editing}

\author[1,2]{Minh-Triet Tran}[orcid=0000-0003-3046-3041] \ead{tmtriet@fit.hcmus.edu.vn}
\credit{Validation, Writing – Review \& Editing}

\author[1,2]{Trung-Nghia Le}[orcid=0000-0002-7363-2610] \ead{ltnghia@fit.hcmus.edu.vn}
\cormark[2]
\credit{Supervision, Conceptualization, Writing – Review \& Editing, Project administration}

\cortext[1]{These authors contributed equally to this work.}
\cortext[2]{Corresponding author}


\affiliation[1]{organization={University of Science}, 
            city={Ho Chi Minh City},
            country={Viet Nam}}

\affiliation[2]{organization={Vietnam National University}, 
            city={Ho Chi Minh City},
            country={Viet Nam}}

\affiliation[3]{organization={Monash University}, 
            state={Victoria},
            country={Australia}}

\affiliation[4]{organization={University of Dayton}, 
            state={Ohio},
            country={United States}}

\begin{abstract}
Customizing image generation remains a core challenge in controllable image synthesis. For single-concept generation, maintaining both identity preservation and prompt alignment is challenging. In multi-concept scenarios, relying solely on a prompt without additional conditions like layout boxes or semantic masks, often leads to identity loss and concept omission. In this paper, we introduce ShowFlow, a comprehensive framework designed to tackle these challenges. We propose ShowFlow-S for single-concept image generation, and ShowFlow-M for handling multiple concepts. ShowFlow-S introduces a KronA-WED adapter, which integrates a Kronecker adapter with weight and embedding decomposition, and together with a novel Semantic-Aware Attention Regularization (SAR) training objective to enhance single-concept generation. Building on this foundation, ShowFlow-M directly reuses robust models learned by ShowFlow-S to support multi-concept generation without extra conditions, incorporating a Subject-Adaptive Matching Attention (SAMA) and a Layout Consistency guidance as the plug-and-play module. Extensive experiments and user studies validate ShowFlow’s effectiveness, highlighting its potential in real-world applications like advertising and virtual dressing. \revise{Our source code will be publicly available at 
\url{https://htrvu.github.io/showflow}.}
\end{abstract}



       



\maketitle

\section{Introduction}\label{sec:introduction}

Personalized text-to-image generation aims to synthesize customized images of specific subjects based on natural language descriptions. Despite the success of large-scale diffusion models~\cite{textualinversion, ruiz2023dreambooth, customdiffusion}, capturing the essence of a user-provided concept while maintaining the model's inherent generative flexibility remains a challenge. This task is bifurcated into two primary objectives: robust \textit{single-concept acquisition} and seamless \textit{multi-concept composition}.

\begin{figure*}[t!]
    \centering
    \includegraphics[width=\textwidth]{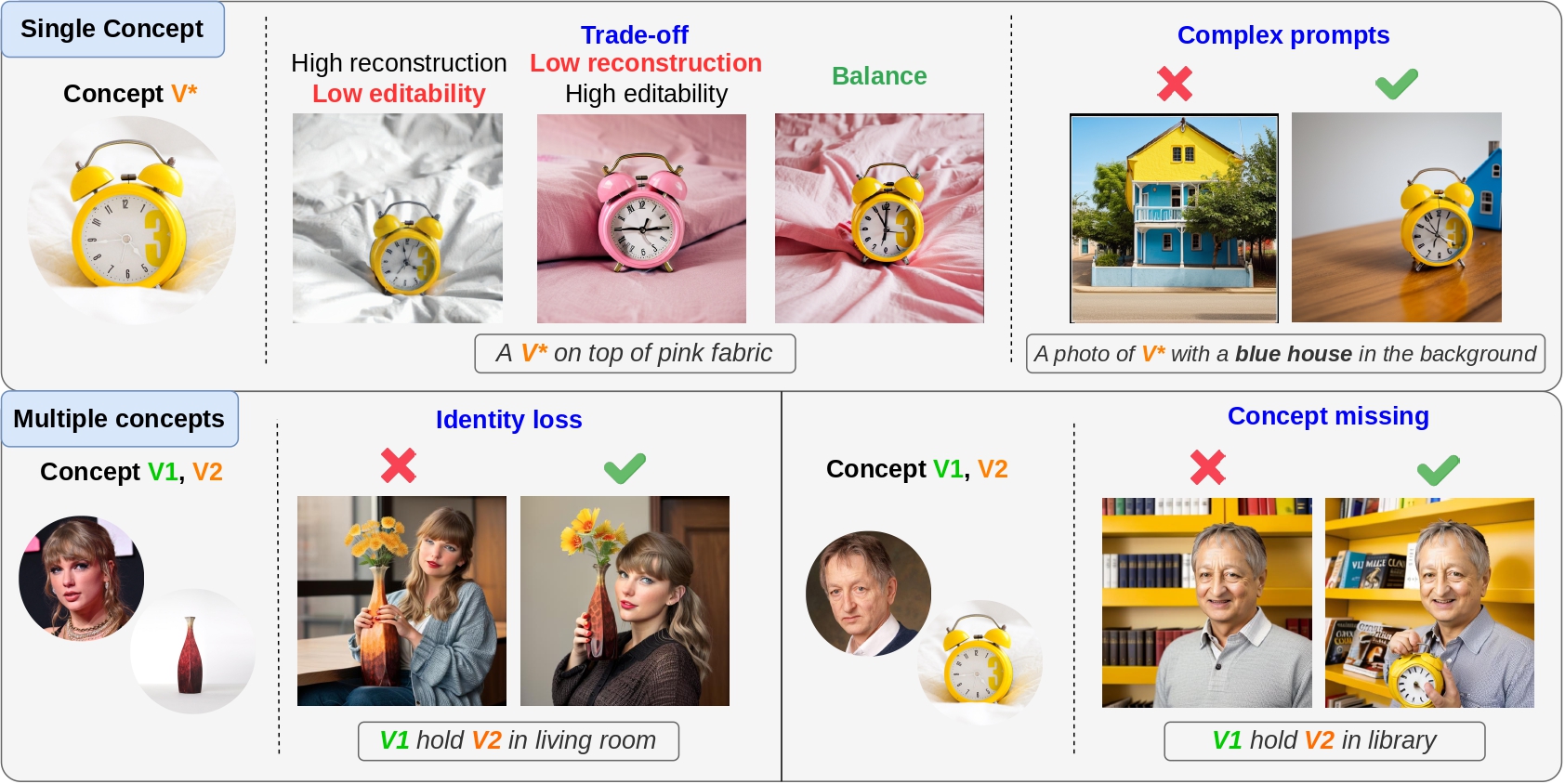}
     \caption{Challenges in personalized image generation. Top: Trade-off between reconstruction and editability, especially in complex prompts in single concept generation. Bottom: Identity loss and concept missing issues in generating multiple concepts.}
    \label{figure:chapter6_intro_challenges}
\end{figure*}

In single-concept generation, the fundamental hurdle is the trade-off between identity reconstruction and prompt-based editability. While full-parameter fine-tuning methods like DreamBooth~\cite{ruiz2023dreambooth} achieve high fidelity, they often suffer from catastrophic forgetting and language drift. Conversely, Parameter-Efficient Fine-Tuning (PEFT) techniques, such as LoRA~\cite{lora, loracommunity}, are inherently constrained by a low-rank assumption. This limited capacity frequently fails to encapsulate intricate subject details, leading to "identity blurring" in complex scenarios as illustrated in Figure~\ref{figure:chapter6_intro_challenges} (top), where the generated images lack the concept of the \textit{yellow clock} or incorrectly associate colors.

To bridge this gap, we introduce the {KronA-WED} adapter within our \textbf{ShowFlow-S} module. By leveraging Kronecker decomposition integrated with weight-decomposed low-rank adaptation, KronA-WED provides a high-rank parameter space that captures fine-grained visual features while maintaining a compact footprint. To further safeguard editability, we propose a novel \textbf{S}emantic-Aware \textbf{A}ttention \textbf{R}egularization {(SAR)} training objective. Unlike prior disentanglement methods~\cite{chendisenbooth,pplus, alaluf2023neural, avrahami2023break} that focus solely on embeddings, our objective refines the cross-attention maps according to the roles of newly introduced concept tokens, ensuring they remain spatially grounded, preventing leakage even in dense prompts.

\begin{figure*}[t!]
      \centering
\includegraphics[width=1.0\textwidth]{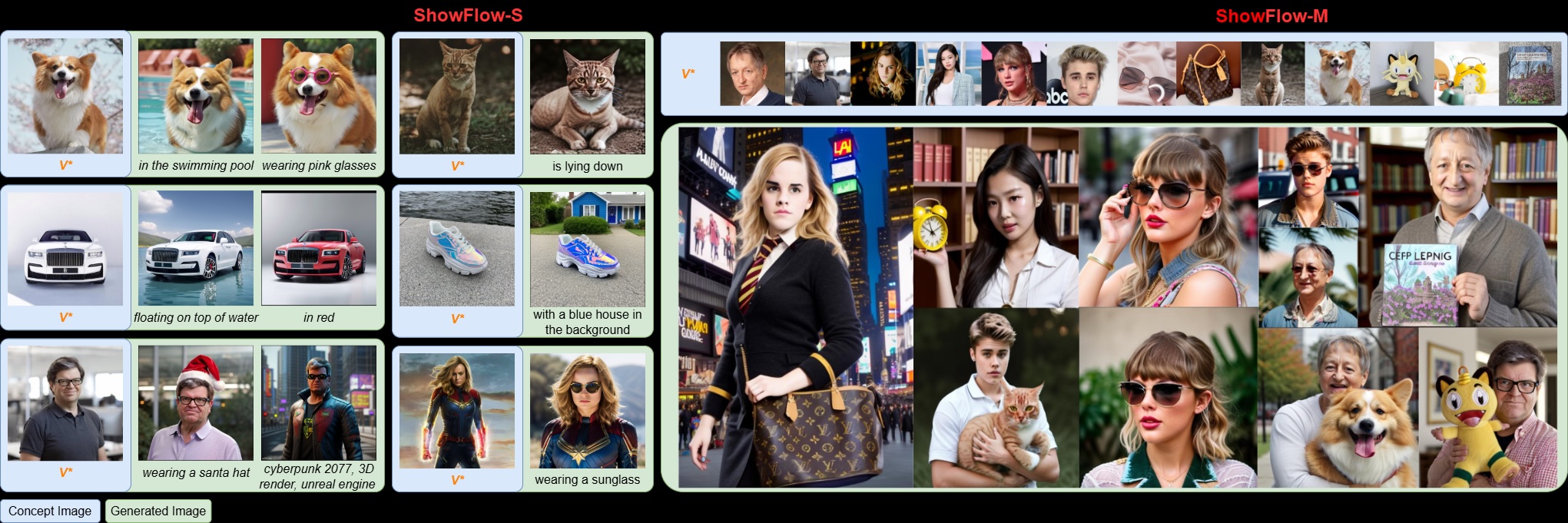}
    \caption{Results of our proposed \textbf{ShowFlow} framework, demonstrating superior customizing image generation across both \textbf{single-concept} and \textbf{multi-concept} scenarios. ShowFlow-S achieves precise identity preservation with prompt alignment, while ShowFlow-M synthesizes complex multi-concept compositions \textit{without requiring additional spatial constraints} such as masks or boxes.}
\label{figure:thesis_chapter6_intro_conceptflow}
\end{figure*}

The challenge intensifies in multi-concept generation, where simply merging models often leads to identity loss or concept omission (Figure~\ref{figure:chapter6_intro_challenges}, bottom). Current state-of-the-art methods~\cite{gu2024mix, kong2024omg, po2024orthogonal, lee2026dreamcatcher, loracomposer} attempt to solve this by requiring external spatial priors like bounding boxes. However, these "layout-dependent" approaches restrict user creativity and struggle with complex interactions. There is a critical need for a "condition-free" framework that can orchestrate multiple subjects purely through text.

To address this, we propose \textbf{ShowFlow-M}, which utilizes a \textbf{S}ubject-\textbf{A}daptive \textbf{M}atching \textbf{A}ttention ({SAMA}) module. SAMA aims to ensure identity preservation where multiple subjects are involved at test-time. This is complemented by our Layout Consistency guidance, which operates in the latent space to ensure that every concept mentioned in the prompt is physically manifested, solving the "missing concept" phenomenon.

As shown in Figure~\ref{figure:thesis_chapter6_intro_conceptflow}, our unified \textbf{ShowFlow} framework demonstrates superior performance across both tasks. \revise{The source code will be publicly available at 
\url{https://htrvu.github.io/showflow}.}

Our contributions are as follows:
\begin{itemize}
    \item We propose {ShowFlow}, a unified, condition-free framework that transitions from robust single-concept acquisition to high-fidelity multi-concept composition without manual spatial guidance.
    \item We introduce the {KronA-WED} adapter and a \textbf{S}emantic-Aware \textbf{A}ttention \textbf{R}egularization {(SAR)} training objective, providing a high-capacity update mechanism that preserves subject identity without sacrificing prompt editability.
    \item We design the \textbf{S}ubject-\textbf{A}daptive \textbf{M}atching \textbf{A}ttention ({SAMA}) module and Layout Consistency guidance, enabling the synthesis of multiple interacting subjects while maintaining distinct identities and preventing concept omission.
    \item Extensive evaluations and user studies demonstrate that ShowFlow significantly outperforms recent baselines in both fidelity and compositional accuracy.
\end{itemize}

\begin{table*}[t!]
\centering
\caption{\revise{Comparison of multi-subject generation methods. Yellow rows highlight methods that share our core setting: layout-free training, layout-free inference, and generalist design principles.}}
\label{tab:method_comparison}
\small 
\revise{
\begin{tabular}{llcccc}
\toprule
\textbf{Method} & \textbf{Venue} & \textbf{\begin{tabular}[c]{@{}c@{}}Layout-Free\\ Inference\end{tabular}} & \textbf{\begin{tabular}[c]{@{}c@{}}Layout-Free\\ Training\end{tabular}} & \textbf{\begin{tabular}[c]{@{}c@{}}Generalist\\ Architecture\end{tabular}} & \textbf{\begin{tabular}[c]{@{}c@{}}Code\\ Released\end{tabular}} \\ 
\midrule
Mix-of-Show~\cite{gu2024mix} & NeurIPS 2023 & \texttimes & \checkmark & \checkmark & \checkmark \\
LoRA-Composer~\cite{yang2025lora} & TIP 2025 & \texttimes & \checkmark & \checkmark & \checkmark \\
MS-Diffusion~\cite{wang2025msdiffusion} & ICLR 2025 & \texttimes & \checkmark & \checkmark & \checkmark \\
DreamCatcher~\cite{lee2026dreamcatcher} & WACV 2026 & \texttimes & \checkmark & \checkmark & \texttimes \\
FastComposer~\cite{xiao2024fastcomposer} & IJCV 2024 & \checkmark & \texttimes & \checkmark & \checkmark \\
MuDI~\cite{jang2024identity} & NeurIPS 2024 & \checkmark & \texttimes & \checkmark & \checkmark \\
TokenVerse~\cite{garibi2025tokenverse} & ACM TOG 2025 & \checkmark & \checkmark & \texttimes & \texttimes \\
Mod-Adapter~\cite{zhong2026modadapter} & ICLR 2026 & \checkmark & \checkmark & \texttimes & \texttimes \\ 
\rowcolor[HTML]{FFFFE0} CustomDiffusion~\cite{customdiffusion} & CVPR 2023 & \checkmark & \checkmark & \checkmark & \checkmark \\
\rowcolor[HTML]{FFFFE0} FreeCustom~\cite{ding2024freecustom} & CVPR 2024 & \checkmark & \checkmark & \checkmark & \checkmark \\
\rowcolor[HTML]{FFFFE0} OMG~\cite{kong2024omg} & ECCV 2024 & \checkmark & \checkmark & \checkmark & \checkmark \\
\midrule
\rowcolor[HTML]{FFFFE0} \textbf{ShowFlow (Ours)} & -- & \checkmark & \checkmark & \checkmark & \checkmark \\ 
\bottomrule
\end{tabular}
}
\end{table*}

\section{Related Work}

\subsection{Single-Concept Generation}

Despite considerable advancements, balancing reconstruction and editability continues to pose a challenge for optimization-based methods~\cite{textualinversion, pplus, ruiz2023dreambooth, customdiffusion}. 
Parameter-efficient fine-tuning methods~\cite{loracommunity, gu2024mix} based on LoRA~\cite{lora} significantly reduce the fine-tuning cost. Still, they often encounter difficulties in capturing complex details of concepts due to the low-rank assumption. To overcome this issue, LyCORIS~\cite{lycoris} adopted Kroncker Adapter (KronA)~\cite{krona} to fine-tune diffusion models. However, only integrating KronA with purely joint weight-embedding tuning methods risks further compromising editability.
DisenBooth~\cite{chendisenbooth} proposed a disentangled learning strategy to improve the editability of joint embedding-weight tuning methods. 
However, attaining effective alignment with complex prompts remains challenging due to wrongly activated regions in token attention maps.




In merging separately learned concepts via gradient fusion~\cite{gu2024mix} for multiple concepts generation, Embedding-Decomposed Low-Rank Adaptation (ED-LoRA)~\cite{gu2024mix}
is the most commonly used adapter. 
Taking it as the baseline, our focus is on improving both the editability and reconstruction capabilities, consequently facilitating the multi-concept generation process.

\subsection{Multiple-Concept Generation}

Recent advancements have pushed the boundaries of customization by attempting to inject multiple novel concepts into a model simultaneously. Early works like Break-a-Scene~\cite{avrahami2023break} and SVDiff~\cite{han2023svdiff} learn individual concepts within multi-subject images but often require ground-truth training data, limiting their practical flexibility. Custom Diffusion~\cite{customdiffusion} achieved multi-concept synthesis through joint optimization, while Mix-of-Show (MoS)~\cite{gu2024mix} introduced gradient fusion to merge separately fine-tuned models. However, these methods often struggle with identity loss or missing concepts when relying solely on text prompts. \revise{To mitigate these issues, recent research has diverged into different streams based on their reliance on spatial layouts and architectural constraints, as detailed in Table~\ref{tab:method_comparison}.}



\revise{A substantial portion of recent methods depends on spatial layouts to orchestrate multiple subjects. MoS~\cite{gu2024mix}, LoRA-Composer~\cite{yang2025lora}, MS-Diffusion~\cite{wang2025msdiffusion}, and DreamCatcher~\cite{lee2026dreamcatcher} are layout-dependent at inference time, meaning they require external spatial priors like bounding boxes or predicted masks to prevent concept bleeding. While effective for simple compositions, these layout-dependent strategies often fail during complex subject interactions or when mask-to-shape mismatches occur, heavily restricting user creativity. Alternatively, another stream of methods, including FastComposer~\cite{xiao2024fastcomposer} and MuDI~\cite{jang2024identity}, achieves layout-free inference but remains layout-dependent at training, requiring layout-driven data during the model's training phase.} 

\revise{In contrast, our work focuses on a strictly \textbf{layout-free setting} during both training and inference. Several methods (\eg, Custom Diffusion~\cite{customdiffusion}, FreeCustom~\cite{ding2024freecustom}, OMG~\cite{kong2024omg}, TokenVerse~\cite{garibi2025tokenverse}, and Mod-Adapter~\cite{zhong2026modadapter}) attempt to achieve this condition-free multi-concept generation without requiring additional spatial conditions at either stage. For example, TokenVerse~\cite{garibi2025tokenverse} and Mod-Adapter~\cite{zhong2026modadapter} effectively address the multi-concept generation problem without spatial priors. However, they lack a generalist design; these approaches are specifically tailored to the modulation of Diffusion Transformers (DiT) and are not compatible with various backbones, resulting in limited broader applicability. Meanwhile, other methods (\eg, OMG~\cite{kong2024omg}) still often resort to spatial conditions in practice because they struggle to orchestrate multiple subjects and prevent concept omissions purely through text. In contrast, our proposed ShowFlow provides a truly generalist, condition-free framework that maintains high identity fidelity and seamlessly synthesizes complex multi-concept interactions purely through text without any spatial guidance.}

\begin{figure*}[t!]
    \centering
    \includegraphics[width=0.7\linewidth]{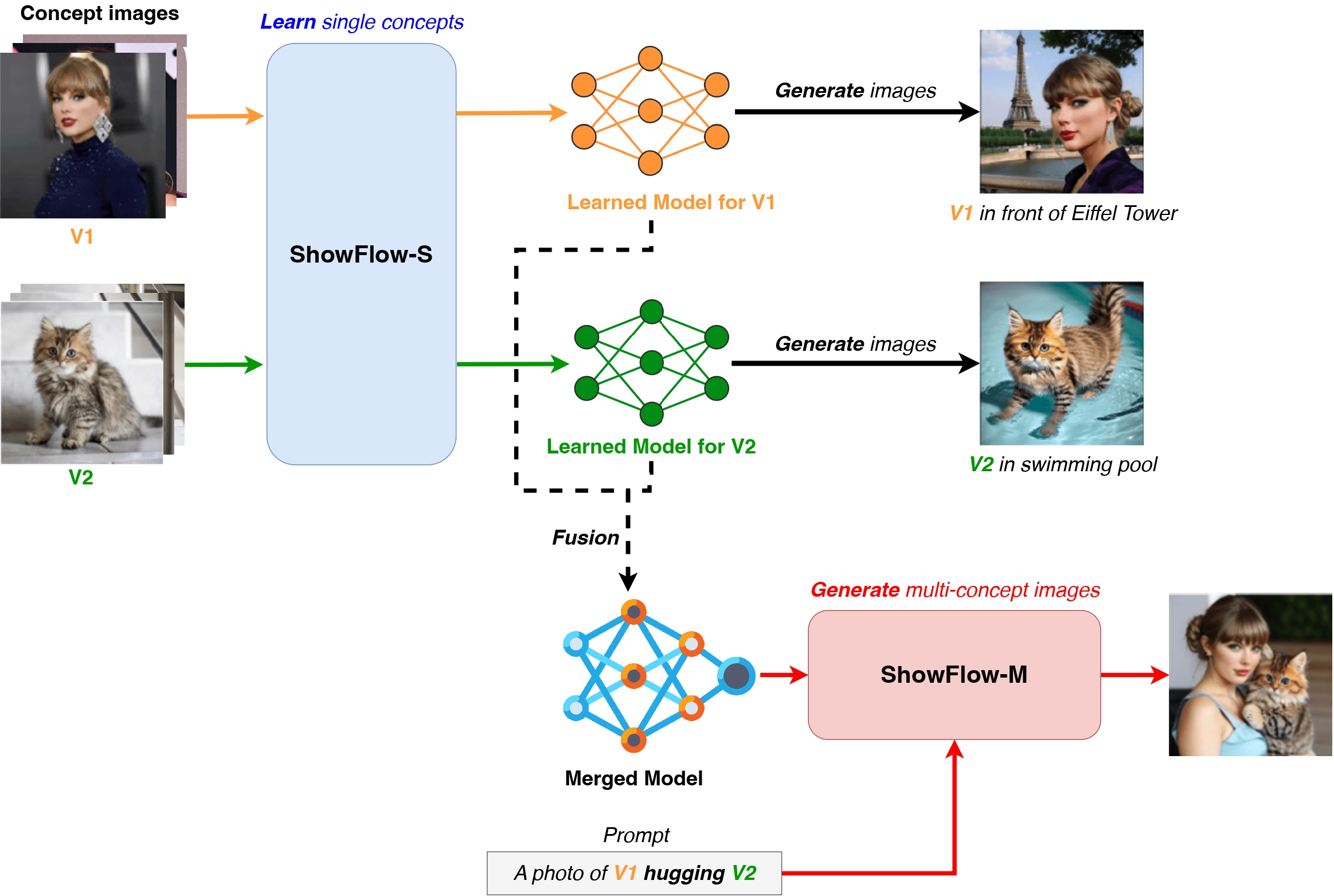}
    \caption{\revise{ShowFlow orchestrates two complementary phases: Stage 1: Robust Concept Acquisition (ShowFlow-S) and Stage 2: Condition-Free Concept Generation (ShowFlow-M).}}
    \label{fig:conceptflow}
\end{figure*}

\section{Preliminaries}
\label{sec:pre}

\paragraph{Parameter-efficient Fine-tuning for Personalization.} Based on the assumption that updates made during the fine-tuning exhibit a low intrinsic rank, Low-Rank Adaptation (LoRA)~\cite{lora} significantly reduces the training parameters by modeling the weight update $\Delta W \in \mathbb{R}^{d \times k}$ using a low-rank decomposition, expressed as $BA$, where $B \in \mathbb{R}^{d \times r}$ and $A \in \mathbb{R}^{r \times k}$, with $r \ll \min(d, k)$. Then, the fine-tuned weight $W'$ can be formulated as:
\begin{equation}
    W' = W_0 + \underline{\Delta W} = W_0 + \underline{BA},
\end{equation}
where $W_0 \in \mathbb{R}^{d \times k}$ is the pre-trained weight matrix and the underlined parameters are being trained during the fine-tuning process. In LoRA, the rank of updated weights is bounded as follow:
\begin{equation}
    \text{rank}(\Delta W) \leq \min(\text{rank}(A), \text{rank}(B)) \leq r.
\end{equation}

Therefore, small $r$ values can impede the ability of LoRA-based methods~\cite{loracommunity, gu2024mix} in capturing complex patterns of concepts in single-concept learning. Increase the $r$ value can enhance this capability but it also leads to a several-fold increase in the model size.

\section{Proposed Method}
\label{sec:method}

\subsection{Problem Statement}
Let $\mathcal{M}_{\theta}$ represent a pre-trained text-to-image diffusion model parameterized by $\theta$. Given a set of $N$ novel concepts $\mathcal{C} = \{C_1, C_2, \dots, C_N\}$, where each concept $C_i$ is defined by a small set of reference images, the overarching goal of personalized multi-concept generation is to synthesize a high-fidelity image $\mathcal{I}$ that seamlessly integrates all $N$ concepts based solely on a joint text prompt $\mathcal{P}$.

A standard approach is to solve the following generation objective:
\begin{equation}
    \mathcal{I} = \mathcal{M}_{\theta + \Delta \theta}(\mathcal{P}, \mathcal{L}),
\end{equation}
where $\Delta \theta$ represents the aggregated parameter updates learned from the individual concept sets, and $\mathcal{L}$ denotes explicit spatial conditions (e.g., bounding boxes or segmentation masks) required to prevent concept blending. 

We identify two fundamental bottlenecks in paradigm:
\begin{enumerate}
    \item \textbf{Capacity vs. Editability Bottleneck in $\Delta \theta$:} Parameter-efficient updates (\eg, low-rank matrices) often lack the representational capacity to memorize high-frequency identity details, while higher-capacity updates risk catastrophic forgetting~\cite{ruiz2023dreambooth}, destroying the model's ability to interpret complex prompts.
    \item \textbf{Layout Dependency ($\mathcal{L}$):} Relying on external spatial conditions $\mathcal{L}$ to resolve attention conflicts restricts user creativity and severely limits the model's ability to generate natural, unconstrained interactions between concepts.
\end{enumerate}

\subsection{Framework Overview}
To achieve high-fidelity, condition-free multi-concept generation, effectively solving for $\mathcal{I} = \mathcal{M}_{\theta + \Delta \theta^*}(\mathcal{P})$ without requiring $\mathcal{L}$, we propose {ShowFlow}, a unified two-stage framework. We hypothesize that successful multi-concept synthesis inherently relies on highly decoupled, overfitting-resistant single-concept representations. Therefore, ShowFlow orchestrates two complementary phases: {Stage 1: Robust Concept Acquisition (ShowFlow-S)} and {Stage 2: Condition-Free Concept Generation (ShowFlow-M)}. \revise{The overall diagram of our proposed framework is shown in Figure~\ref{fig:conceptflow}.}

In Stage 1, we address the capacity-editability bottleneck by optimizing the model's parameter space to memorize high-frequency identity details using our KronA-WED adapter, coupled with a Semantic-Aware Attention Regularization (SAR) training objective. In Stage 2, we eliminate layout dependency by freezing these robust weights and shifting our focus to the activation space. We introduce the Subject-Adaptive Matching Attention (SAMA) module and Layout Consistency guidance to dynamically route visual features and compose multiple concepts seamlessly without external spatial constraints.

\subsection{Robust Concept Acquisition (ShowFlow-S)}
\label{sec:singleconcept}

\begin{figure*}[t!]
    \centering
    \includegraphics[width=\linewidth]{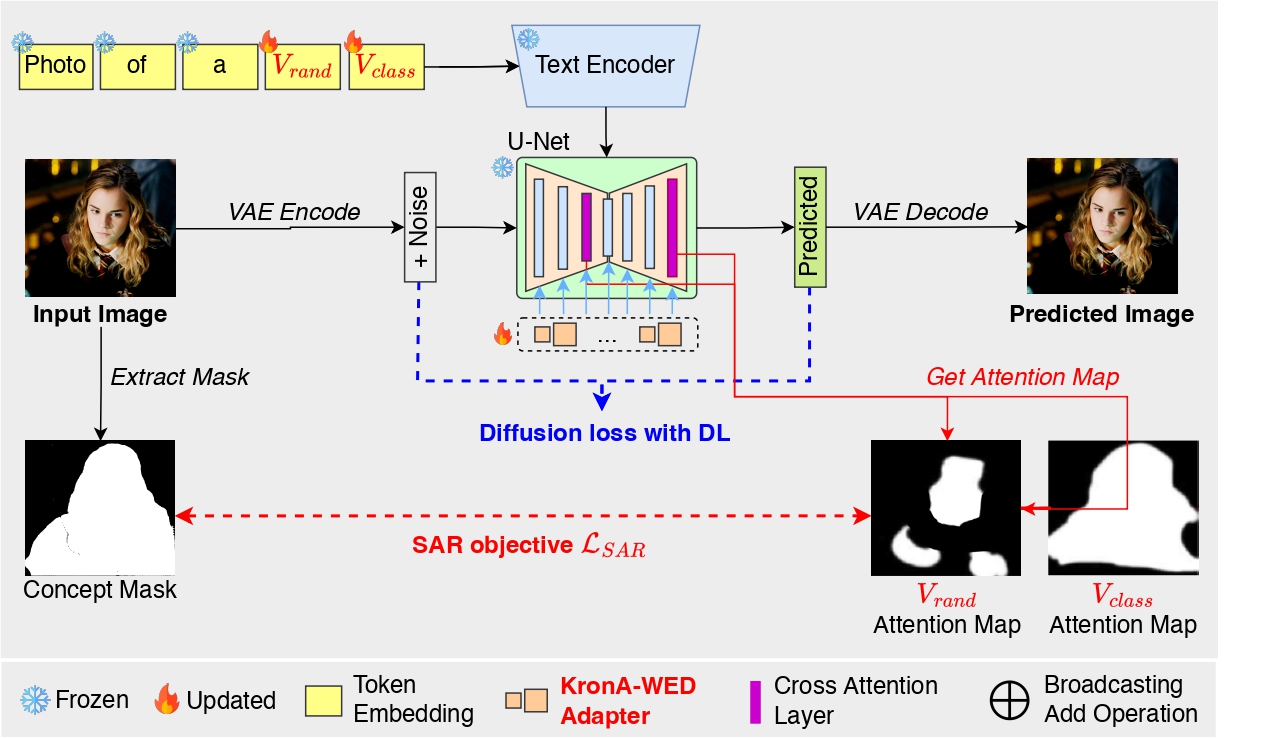}
     \caption{Pipeline of ShowFlow-S for single-concept learning. We fine-tune the newly-added tokens ($V_{rand}$ and $V_{class}$) and our proposed KronA-WED adapters. The training objectives include our proposed Semantic-Aware Attention Regularization (SAR) loss $\mathcal{L}_{SAR}$ and diffusion loss with disentangled learning (DL)~\cite{chendisenbooth} to further enhance editability.}
    \label{figure:chapter6_single_proposed}
    \vspace{-0.2cm}
\end{figure*}

The goal of this stage is to learn a new concept from a few user-provided images while preventing language drift. Taking ED-LoRA~\cite{gu2024mix} as a baseline, we introduce the KronA-WED adapter and a fine-tuning strategy that integrates disentangled learning~\cite{chendisenbooth} with our proposed Semantic-Aware Attention Regularization (SAR) training objective. The pipeline for ShowFlow-S is depicted in Figure~\ref{figure:chapter6_single_proposed}.

\subsubsection{Enhanced Identity Learning via KronA-WED}
\begin{figure*}[t!]
    \centering
    \includegraphics[width=\linewidth]{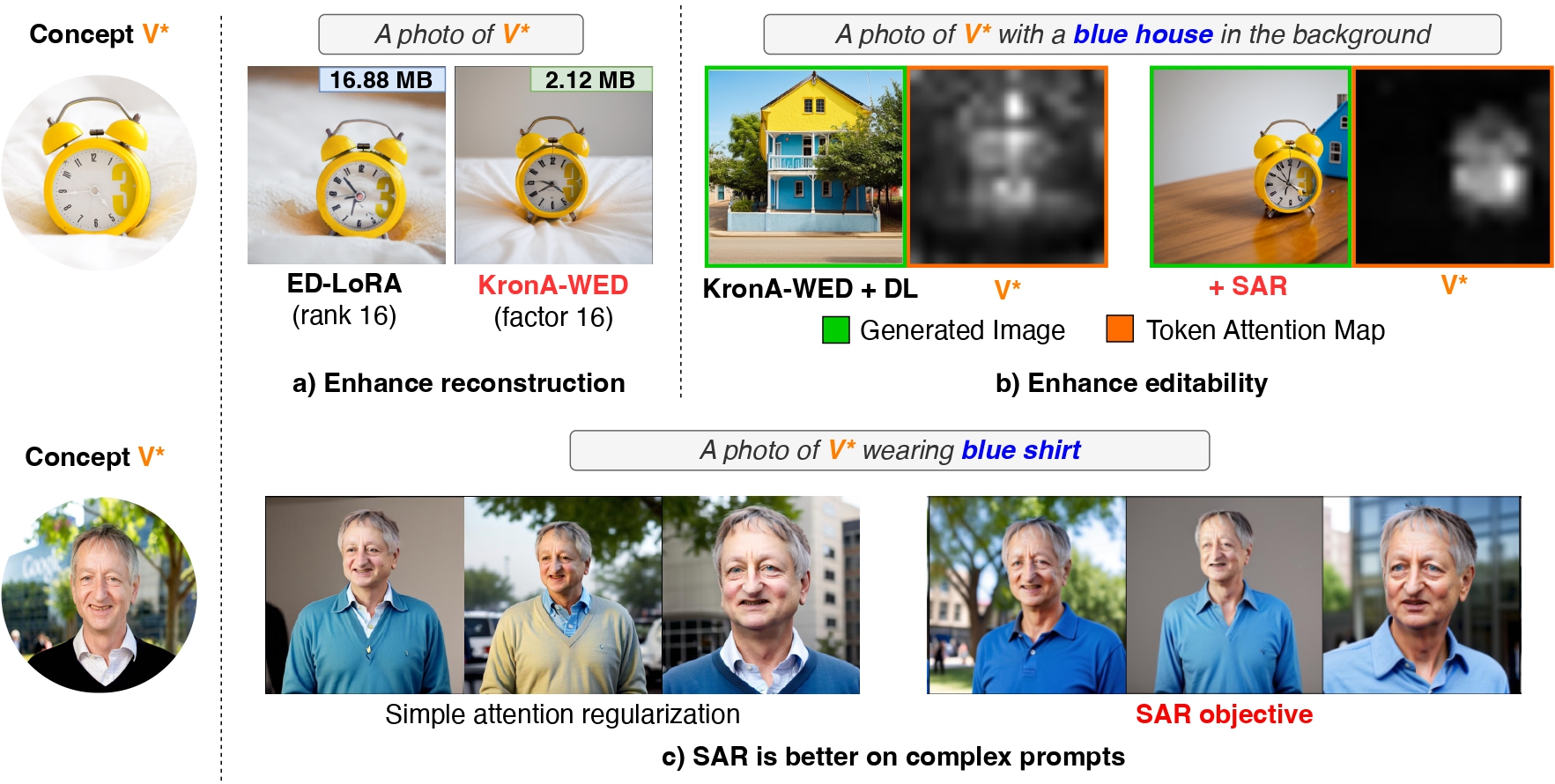}
     \caption{ShowFlow-S balances the trade-off. (a) The KronA-WED adapter captures complex concept patterns while maintaining a compact model size. (b) Incorporating Semantic-Aware Attention Regularization (SAR) into disentangled learning (DL)~\cite{chendisenbooth} improves prompt alignment under complex prompts. (c) SAR outperforms the simple attention regularization used in prior work~\cite{avrahami2023break, chefer2023attend} when handling complex prompts.}
    \label{figure:chapter6_single_effectiveness}
\end{figure*}

Standard Low-Rank Adaptation (LoRA) often lacks the capacity to capture intricate textures required for high-fidelity personalization, as the low-rank assumption impedes the ability to learn complex patterns (see Figure~\ref{figure:chapter6_single_effectiveness}a). To increase the parameter rank without inflating the computational footprint, we introduce the \textbf{Kron}ecker \textbf{A}daptation with \textbf{W}eight and \textbf{E}mbedding \textbf{D}ecomposition ({KronA-WED}) module. 

We extend the baseline architecture by replacing standard LoRA modules with Kronecker Adapters (KronA)~\cite{krona}, where the updated weight $\Delta W \in \mathbb{R}^{d \times k}$ is expressed as $A \otimes B$, where $A \in \mathbb{R}^{a_1 \times a_2}$ and $B \in \mathbb{R}^{b_1 \times b_2}$, with $a_1 \times b_1 = d$ and $a_2 \times b_2 = k$. Under this decomposition, we have
\begin{equation}
    \text{rank}(\Delta W) = \text{rank}(A) \cdot \text{rank}(B) \leq \min(a_1, a_2) \cdot \min \left ( \frac{d}{a_1}, \frac{k}{a_2} \right ).
\end{equation}
Therefore, we can increases the effective update rank beyond LoRA, approaching the capacity of full fine-tuning while maintaining parameter efficiency. 
However, naively injecting unconstrained high-rank updates risks disrupting the model's pre-trained structural priors, often leading to training instability. We hypothesize that subject-specific personalization should primarily act as a directional shift in the feature space rather than a magnitude distortion. To enforce this, we integrate the high-capacity Kronecker adaptation with a decoupled optimization strategy. Building upon the directional update principles introduced in DoRA~\cite{dora}, we explicitly decouple the network weights into learnable magnitude and directional components, confining the Kronecker update strictly to the directional space. The resulting KronA-WED fine-tuned weights $W'$ are formulated as:

\begin{equation}
    W' = \underline{m} \frac{W_0 + \underline{\Delta W}}{||W_0 + \underline{\Delta W}||_c} = \underline{m} \frac{W_0 + \underline{A \otimes B}}{||W_0 + \underline{A \otimes B}||_c},
    \label{equation:chapter6_single_krona-dora}
\end{equation}
where $W_0$ is the frozen pre-trained weight, $A$ and $B$ are the trainable Kronecker factors, $m$ is a learnable magnitude vector, and $||\cdot||_c$ is the vector-wise norm across each column. We use He initialization~\cite{he2015delving} for $A$ and zero for $B$, while $m$ is initially set to $||W_0||_c$. For the KronA decomposition, we use a single factor $f$~\cite{lycoris} that is close to $\sqrt{d}$ and $\sqrt{k}$ value. As shown in Figure~\ref{figure:chapter6_single_effectiveness}a, this formulation allows the model to make high-capacity directional updates to capture fine-grained subject details deeply within the network weights.

\subsubsection{Semantic-Aware Attention Regularization for Better Editability} 
While disentangled learning~\cite{chendisenbooth} enhances general editability by learning separate textual and visual embeddings to prevent the mixing of a concept's identity with irrelevant background details, without proper supervision, the cross-attention maps of newly added tokens may become overly spread out and activate incorrect regions, disrupting alignment in highly intricate prompts (Figure~\ref{figure:chapter6_single_effectiveness}b). 

To ensure concept tokens remain spatially grounded, we propose a novel \textbf{S}emantic-Aware \textbf{A}ttention \textbf{R}egularization \textbf{(SAR)} training objective. Specifically, we represent the concept using two new tokens: $V_{rand}$ (adjective) and $V_{class}$ (noun). Intuitively, the noun token $V_{class}$ is encouraged to align tightly with the foreground mask of the concept, as its cross-attention map should activate the spatial region where the concept appears. In contrast, the adjective token $V_{rand}$ focuses on attribute-specific regions and therefore only requires partial activation within the foreground mask, allowing it to capture distinctive details of the concept such as the face, hair, texture, or other characteristic attributes. 

For each training image $\mathbf{x}_i$, we extract a foreground mask $M_i$ using a background removal model. We penalize the deviation of the cross-attention maps from these masks as follow:
\begin{equation}
\begin{aligned}
        \mathcal{L}_{SAR} = \lambda_{SAR}\sum_{i=1}^{N} \frac{1}{2} \Big( & || C_{\theta}(V_{rand}, \mathbf{z}_{i, t_i}) \odot (1 - M_i) ||_F^2 + \\ 
        & || C_{\theta}(V_{class}, \mathbf{z}_{i, t_i}) \odot M_i ||_F^2 \Big),
\end{aligned}
\end{equation}
where $C_{\theta}(V, \mathbf{z}_{i, t_i})$ represents the average cross-attention map between token $V$ and the noisy latent $\mathbf{z}_{i, t_i}$ at timestep $t_i$, $\odot$ is the Hadamard product, and $||\cdot||_F$ is the Frobenius norm. 
We empirically set the regularization weight $\lambda_{SAR} = 0.001$ to balance the gradient magnitudes between the diffusion loss and the spatial penalty, preventing identity collapse while maintaining editability. As shown in Figure~\ref{figure:chapter6_single_effectiveness}c, compared to the simple attention regularization used in prior work~\cite{avrahami2023break, chefer2023attend}, which applies the entire foreground mask as the supervision signal for the cross-attention map of all concept tokens, our SAR objective improves the robustness of the model to complex prompts (\eg, interaction between objects).

\subsection{Condition-Free Concept Generation (ShowFlow-M)}
\label{sec:multiconcept}

\begin{figure*}[t!]
  \centering
  \includegraphics[width=\linewidth]{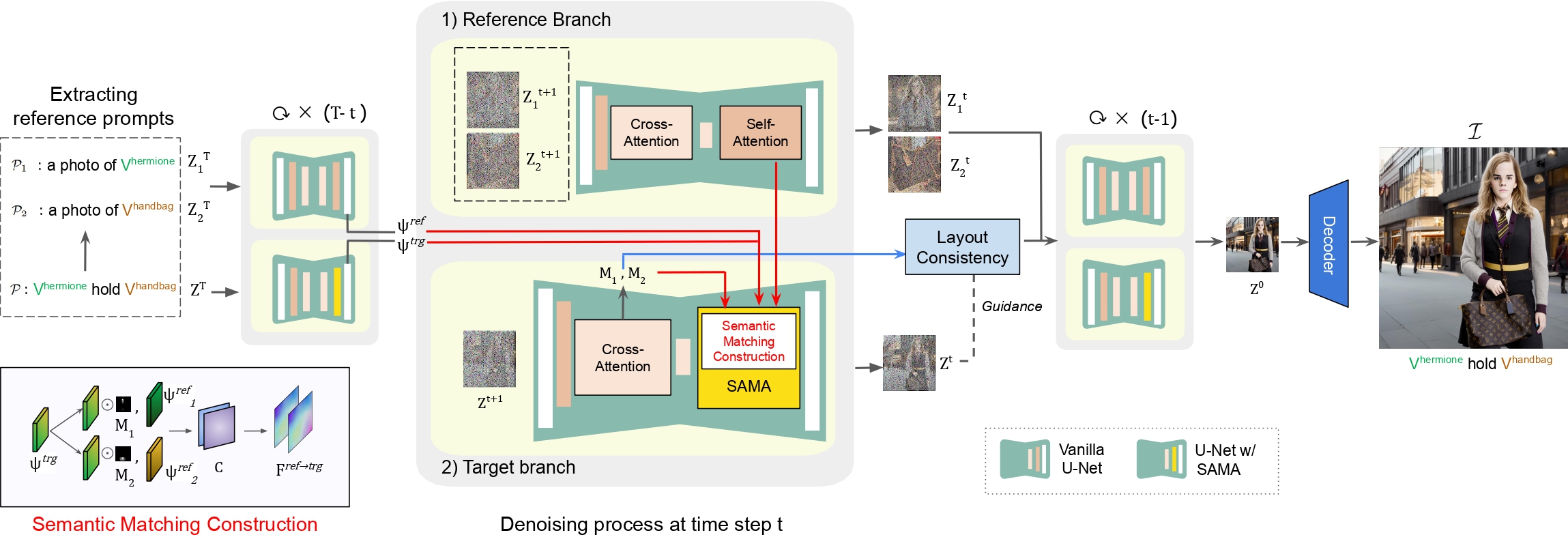}
  \caption{Overview of the composite pipeline in ShowFlow-M for two concepts. The denoising process features two branches: reference and target. We extract concept tokens in the target prompt \(\mathcal{P}\) to form reference paths. At each timestep, features from the reference branch enhance the identity details in the target branch through our Subject-Adaptive Matching Attention (SAMA) module. The latent is continuously updated by Layout Consistency guidance.}
  \label{fig:chapter6_overall_multiconcept}
  \vspace{-0.2cm}
\end{figure*}

Given $N$ independently learned KronA-WED concepts $\{\Delta\theta_n\}_{n=1}^{N}$ and their associated tokens $\{S^*_n\}_{n=1}^{N}$ (where $S^*_n = V_{rand}^n V_{class}^n$) from Stage 1, ShowFlow-M fuses them into a single generation process. We aggregate the weights using gradient fusion~\cite{gu2024mix} to obtain $\epsilon_{\theta_0 + \Delta\theta}$. However, merely combining weights inevitably leads to concept bleeding because standard cross-attention cannot dynamically disambiguate multiple personalized tokens. To resolve this without requiring external spatial constraints (\eg, bounding boxes), we introduce Subject-Adaptive Matching Attention (SAMA) module and Layout Consistency guidance. The composite pipeline is illustrated in Figure~\ref{fig:chapter6_overall_multiconcept}.

\subsubsection{Subject-Adaptive Matching Attention (SAMA)}
\label{sec:sa}

While fused models render individual appearances well, they struggle with identity fidelity when multiple concepts interact. {SAMA} solves this by enhancing multi-concept identity preservation in the \textit{target} branch using routing from single-concept \textit{reference} branches during denoising. For each concept, we identify semantically similar features between the reference and target branches. Based on this correspondence, we \textit{warp} the reference features to spatially align them with those in the target branch, a process we refer to as \textbf{semantic matching construction}, thereby incorporating the aligned features into the target representation. We apply SAMA to the \textit{self-attention} layers of the middle block and earlier blocks of the decoder in U-Net as the spatial features there have significant semantic appearance information~\cite{tumanyan2023plug,zhang2024tale,mou2024dragondiffusion}.

Specifically, for a target prompt $\mathcal{P}$ containing $K$ concepts, we generate $K$ parallel reference branches, each guided by a simple prompt ``\textit{a photo of} \(S^\mathcal{P}_k\)". At each timestep, we extract intermediate U-Net layer feature descriptors $\{\psi^{ref}_k\}_{k=1}^{K}$ from the reference branches and $\psi^{trg}$ from the target branch. To explicitly correlate these features, we compute a dense cost volume $\mathbf{C}_k$ for each concept:
\begin{equation}
\mathbf{C}_k(x, y) = \frac{{\psi^{trg} \odot \mathbf{M}_k}(x) \cdot \psi^{ref}_k(y)}{\|\psi^{trg} \odot \mathbf{M}_k(x)\| \|\psi^{ref}_k(y)\|},
\end{equation}
where $\mathbf{M}_k$ is a foreground mask computed from the token's cross-attention map of this concept. We then compute the semantic flow $\mathbf{F}_k^{ref \rightarrow trg}$ via argmax over $\mathbf{C}_k$, and use it to warp the reference value features:
\begin{equation}
\mathbf{V}_k^{ref \rightarrow trg} = \mathcal{W}(\mathbf{V}_k^{ref}; \mathbf{F}_k^{ref \rightarrow trg}) \odot \mathbf{M}_k.
\end{equation} 

\begin{figure}[t!]
	\centering
	\includegraphics[width=\linewidth]{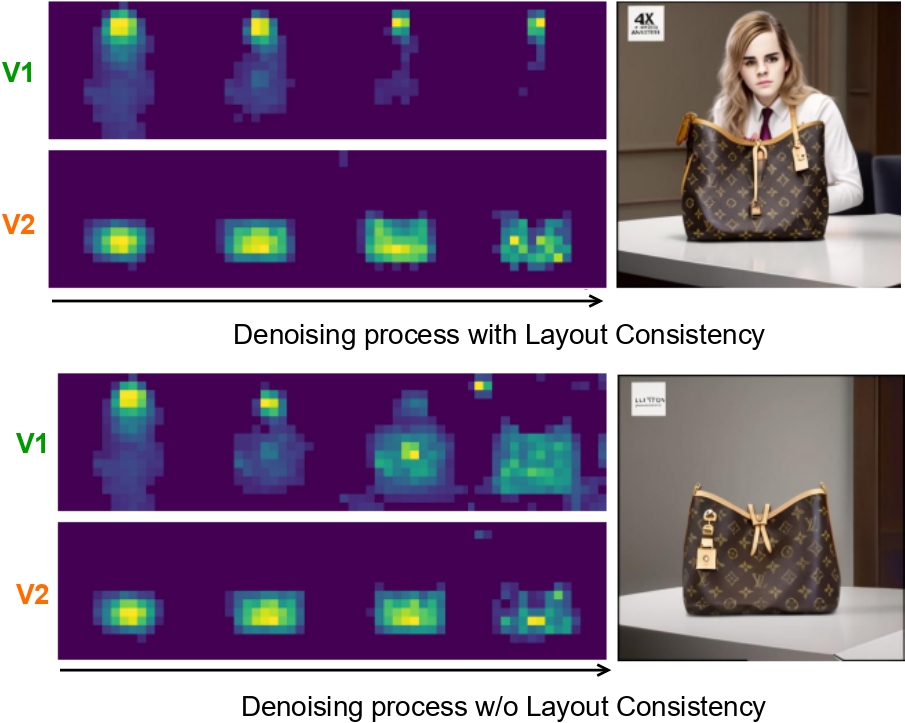}
	\caption{Illustration of layout consistency guidance with the prompt \textit{``$V_1$ advertising $V_2$''}. Ensuring consistency between successive attention maps helps maintain the intended layout of concepts and prevents concept omission.}
	\label{fig:chapter6_layout_loss}
\end{figure}

\begin{figure*}[t!]
    \centering
    \includegraphics[width=\linewidth]{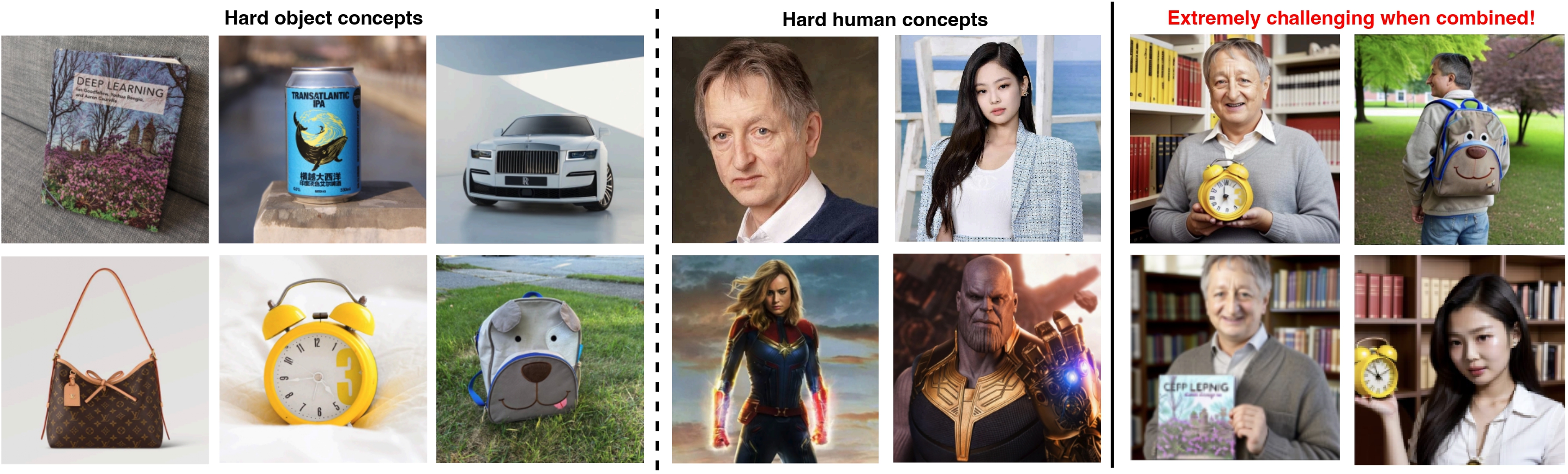}
    \caption{\revise{Examples of challenging concepts in our dataset. When composed together in multi-concept generation, these cases significantly increase the difficulty of preserving identity and maintaining compositional fidelity.}}
    \label{fig:rebuttal_hard_concepts}
\end{figure*}

To maintain spatial consistency across the background and all concepts, we aggregate the warped features:
\begin{equation}
\mathbf{V}^{\mathcal{W}} = \sum_{k=1}^{K} \mathbf{V}_k^{ref \rightarrow trg} + \mathbf{V}^{trg} \odot \left(1 - \sum_{k=1}^{K} \mathbf{M}_k\right).
\end{equation}
SAMA then integrates \(\mathbf{V}^{\mathcal{W}}\) directly into the self-attention module of the target branch:
\begin{equation}
\label{equ:sama_attention}
\begin{aligned}
\mathrm{SAMA}(\mathbf{Q}^{trg},\mathbf{K}^{trg}, \mathbf{V}^{\mathcal{W}})= \mathrm{Softmax}\left(\frac{\mathbf{Q}^{trg} (\mathbf{K}^{trg})^T}{\sqrt{d}}\right)\mathbf{V}^{\mathcal{W}}.
\end{aligned}
\end{equation}
By restricting attention pathways at the activation level, SAMA prevents attribute leakage across distinct subjects.

\subsubsection{Layout Consistency Guidance}
\label{sec:ca}

Even with optimal feature routing, generating multiple concepts often suffers from the "missing concept" problem due to layout deterioration during the denoising process~\cite{agarwal2023star}. As shown in Figure~\ref{fig:chapter6_layout_loss}, a layout may initially include all key concepts at $t = T$ but misalign by $t = 0$.

To solve this condition-free, we propose a test-time latent Layout Consistency guidance. We maximize the Intersection over Union (IoU) of activated regions across timesteps to anchor the semantics. Let \(\mathbf{A}_k^t(i,j)\) represent the refined activation map derived from the cross-attention map \(\mathbf{M}_k^t(i,j)\):
\begin{equation}
\mathbf{A}_k^t(i,j) =
\begin{cases}
\mathbf{M}_k^t(i,j) + \lambda, & \text{if } \mathbf{M}_k^t(i,j) > \tau, \\
\mathbf{M}_k^t(i,j) - \lambda, & \text{if } \mathbf{M}_k^t(i,j) \leq \tau,
\end{cases}
\end{equation}
where \(\lambda\) is an adjustment factor and \(\tau\) is a threshold. The layout Consistency loss is defined as:
\begin{equation}
\label{equ:attention_guidance}
\mathcal{L}_{\text{Layout}} = \sum_{k=1}^K \left( 1 - \frac{\sum_{i,j} \mathbf{A}_k^t(i,j)  \mathbf{A}_k^{T}(i,j)}{\sum_{i,j} \max(\mathbf{A}_k^t(i,j), \mathbf{A}_k^{T}(i,j))} \right).
\end{equation}
At each denoising step, we apply a targeted energy gradient to the latent code \(\mathbf{z}_t\), scaled by a linearly decreasing decay factor \(\phi_t\):
\begin{equation}
\label{equ:latent_update}
\mathbf{z}_t^{\prime} = \mathbf{z}_t - \phi_t \cdot \nabla_{\mathbf{z}_t} \mathcal{L}_{\text{Layout}}.
\end{equation}
This latent operation ensures that all requested entities physically materialize in the final image, circumventing the need for user-provided bounding boxes.

\begin{figure}[t!]
	\centering
	\includegraphics[width=\linewidth]{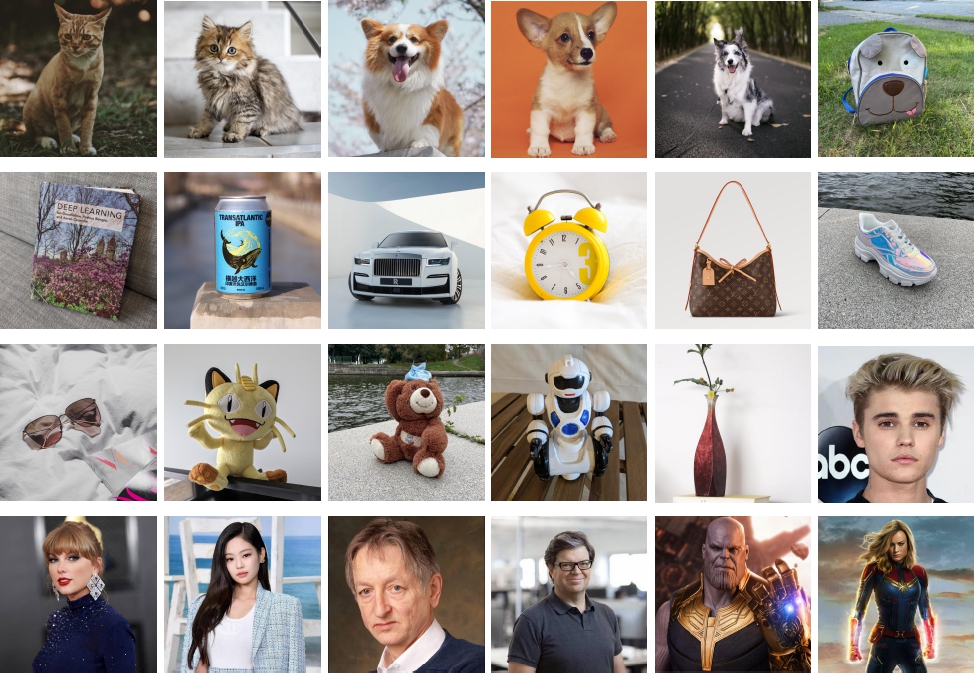}
	\caption{All concepts in our dataset for experiments.}
	\label{figure:chapter6_experiment_dataset}
\end{figure}

\section{Experiments}

\subsection{Experimental Settings}

\vspace{0.15cm}
\noindent
\textbf{Datasets.} While the DreamBench~\cite{ruiz2023dreambooth} and CustomConcept101~\cite{customdiffusion} datasets contain only salient objects and animals, we extend our benchmark with character concepts to enable a more comprehensive evaluation in both single and multiple concept settings.

Our dataset includes 12 objects, 5 animals, and 7 characters (5 \textit{regular humans} and 2 \textit{special humans}), which are shown in Figure~\ref{figure:chapter6_experiment_dataset}. 
The key distinction between regular humans and special humans is that for the latter, we aim to also preserve their outfits, while for regular human characters, our primary focus is only on their faces. For multiple concepts generation, we focus on the interaction between character-object and character-animal, resulting in 60 combinations.

\revise{In our dataset, we include challenging concepts from DreamBench~\cite{ruiz2023dreambooth}, such as objects with fine-grained details (\eg small texts and intricate visual patterns), as well as human subjects with distinctive attributes (\eg unique outfits and accessories). These design choices introduce significant challenges in preserving identity and fine details, particularly in multi-concept settings. Moreover, when these challenging concepts are combined, the task becomes substantially more difficult due to increased competition in attention and representation, making identity preservation and compositional accuracy more demanding (See Figure~\ref{fig:rebuttal_hard_concepts}).}

\begin{figure*}[t!]
    \centering

    \includegraphics[width=0.95\linewidth]{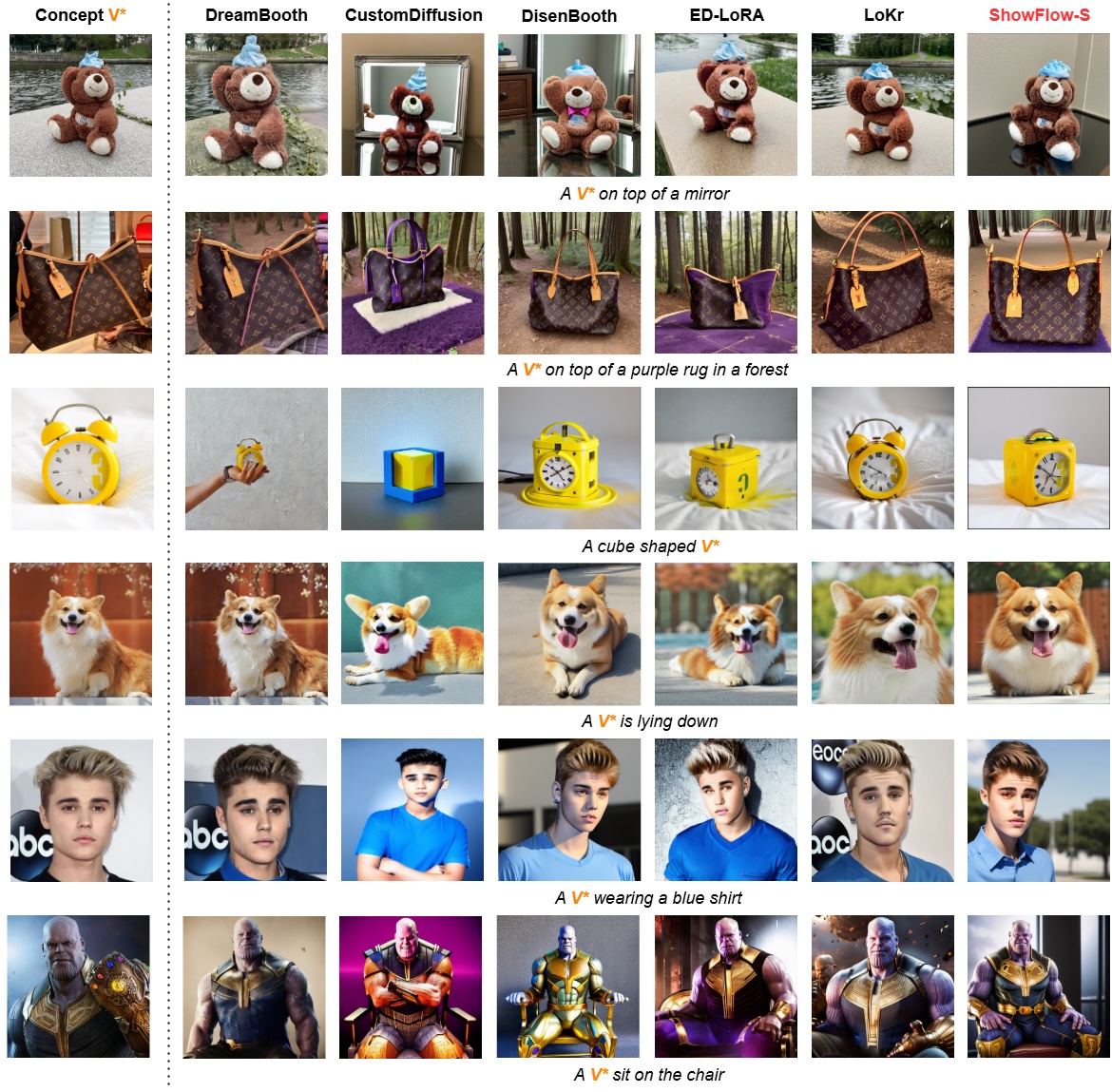}

    
     \caption{Qualitative comparison on single concept generation between ShowFlow-S and other baselines. We focus on the evaluation with complex prompts.}
    \label{figure:chapter6_single_qualitative}
    
    \vspace{-0.25cm}
\end{figure*}

\begin{figure*}[t!]
    \centering

    \includegraphics[width=0.9\linewidth]{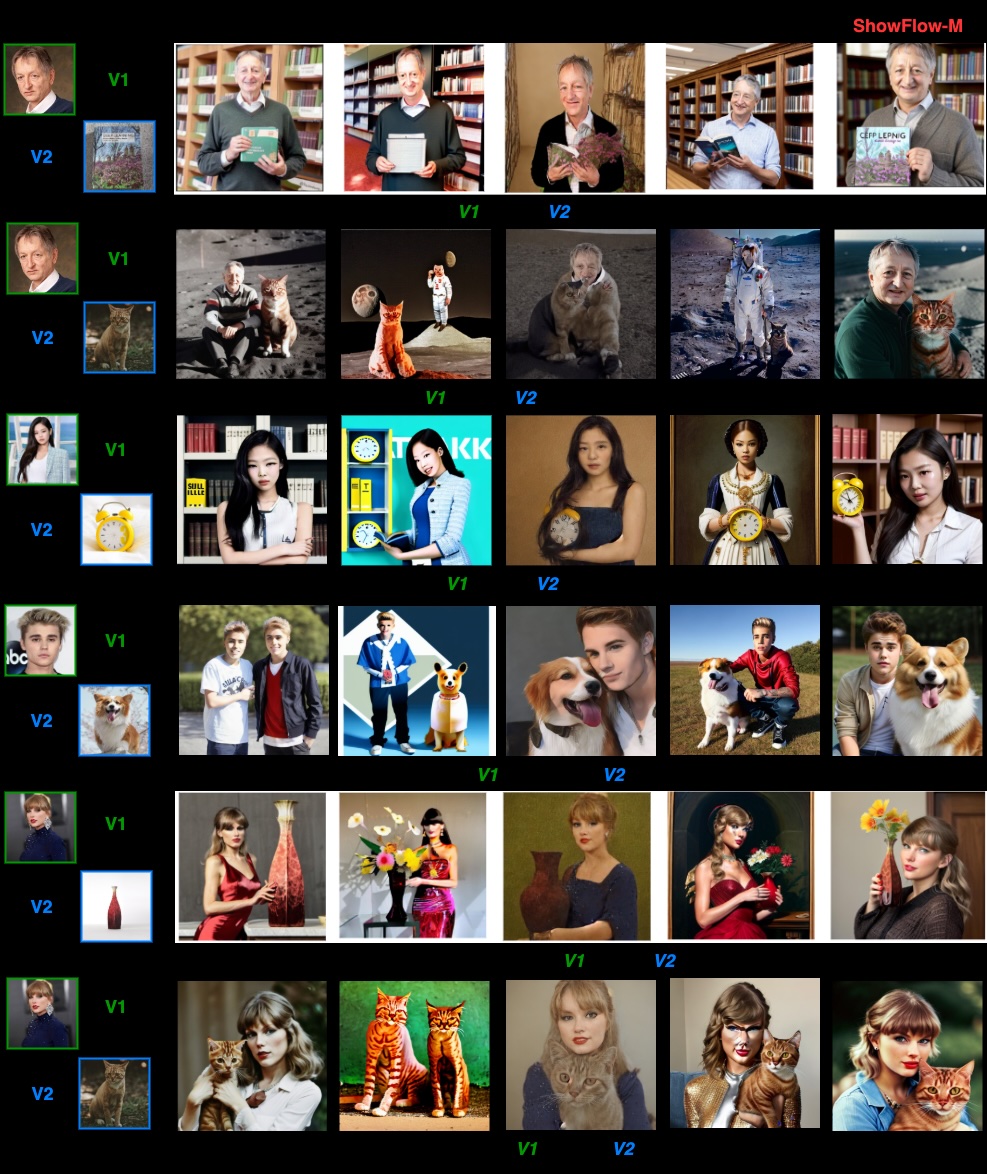}
    
     \caption{Qualitative comparison between ShowFlow-M and other baselines. 
     }
  \vspace{-10pt}
     
    \label{figure:chapter6_multi_qual_baseline}
\end{figure*}

\vspace{0.2cm}
\noindent
\textbf{Implementation Details.} For single concept learning, we incorporate KronA-WED to all linear layers in all attention modules of the U-Net with the decomposition factor $f=16$. 
We set the weight $\lambda_{SAR} = 0.001$ while the other weights are based on disentangled learning settings~\cite{chendisenbooth}.
The unified weight $\Delta\theta$ for ShowFlow-M is fused from the two weights learned by ShowFlow-S with gradient fusion~\cite{gu2024mix}.



\vspace{0.2cm}
\noindent
\textbf{Evaluation Setting.}
The prompts for single concept generation are divided into four main types, including 
Recontextualization, Restylization, Interaction, and Property Modification, and prompt templates are borrowed from previous works~\cite{ruiz2023dreambooth, gu2024mix}.
For multiple concepts generation, we define a list of prompts for each combination that focuses on the interaction between concepts, especially character-object and character-animal interactions.

\vspace{0.2cm}
\noindent
\textbf{Evaluation Metrics.}
To evaluate the identity preservation capability, we adopt the ViT-S/16 DINO score proposed by~\cite{ruiz2023dreambooth}, \ie, the average pairwise cosine similarity between the DINO~\cite{caron2021emerging} embeddings of the generated images and the input real images. The prompt alignment is evaluated by the average cosine similarity between the text prompt and image CLIP~\cite{clip} embeddings.
It is noteworthy that we calculate the DINO score separately for each concept in multi-concept images and then average them to get the final result. 
Additionally, we scale the CLIP-T score by a factor of 2.5 to match the range of DINO, and then compute the F1 score between the two as a provisional metric to assess the balance between reconstruction and editability.
To further evaluate identity preservation in human-related concepts, we incorporate the ArcFace~\cite{deng2019arcface} metric for all methods in both tasks.

\vspace{0.2cm}
\noindent
\textbf{Baselines.}
We compare ShowFlow-S with other jointly embedding and weight tuning methods, including DreamBooth~\cite{ruiz2023dreambooth}, Custom Diffusion~\cite{customdiffusion}, DisenBooth~\cite{chendisenbooth} ED-LoRA~\cite{gu2024mix}, and LoKr module from LyCORIS~\cite{lycoris}. 
For ShowFlow-M, we compare our method with Mix-of-Show~\cite{gu2024mix}, CustomDiffusion~\cite{customdiffusion}, FreeCustom~\cite{ding2024freecustom}, and OMG~\cite{kong2024omg} in a condition-free setting. 


\vspace{1cm}
\subsection{Comparison with Baselines}
\label{subsubsection:chapter6_single_comparison}

\begin{table*}[t!]
\centering
\caption{Quantitative comparison between ShowFlow and other baselines. \textcolor{red}{\textbf{Bold}}, \textcolor{blue}{\underline{underline}}, and \textcolor{olive}{\textit{italics}} indicate the top 1, top 2, and top 3 scores, respectively.}

\begin{subtable}[t]{.5\linewidth}
    \centering
    \scriptsize
    \caption{Single concept generation.}
    \label{table:chapter6_exps_single_quant}
    \begin{tabular}{lcccc}
        \toprule
        \textbf{Methods} & \textbf{DINO$\uparrow$} & \textbf{CLIP-T$\uparrow$} & \textbf{F1-Score$\uparrow$} & \textbf{ArcFace$\uparrow$} \\ 
        \midrule
        DreamBooth~\cite{ruiz2023dreambooth} & \textcolor{red}{\textbf{0.684}} & 0.271 & 0.388 & 0.305 \\
        CustomDiffusion~\cite{customdiffusion} & 0.503 & \textcolor{red}{\textbf{0.313}} & 0.613 & 0.172 \\
        DisenBooth~\cite{chendisenbooth} & 0.616 & \textcolor{blue}{\underline{0.297}} & 0.674 & 0.257 \\
        ED-LoRA~\cite{gu2024mix} & 0.667 & 0.281 & \textcolor{blue}{\underline{0.685}} & \textcolor{olive}{\textit{0.370}} \\
        LoKr~\cite{lycoris} & \textcolor{olive}{\textit{0.679}} & 0.275 & \textcolor{olive}{\textit{0.683}} & \textcolor{blue}{\underline{0.375}} \\
        \midrule
        \rowcolor[HTML]{FFFFE0} \textbf{ShowFlow-S} & \textcolor{blue}{\underline{0.682}} & \textcolor{olive}{\textit{0.282}} & \textcolor{red}{\textbf{0.694}} & \textcolor{red}{\textbf{0.397}} \\
        \bottomrule
    \end{tabular}
        
\end{subtable}%
\begin{subtable}[t]{.5\linewidth}
    \centering
    \scriptsize
    \caption{Multiple concepts generation.}
    \label{table:chapter6_multi_quan_compare}
    \begin{tabular}{lcccc}
        \toprule
            \textbf{Methods} & \textbf{DINO$\uparrow$} & \textbf{CLIP-T$\uparrow$} & \textbf{F1-Score$\uparrow$} & \textbf{ArcFace$\uparrow$} \\ 
            \midrule
            Mix-of-Show~\cite{gu2024mix} & \textcolor{blue}{\underline{0.436}} & \textcolor{olive}{\textit{0.312}} & \textcolor{blue}{\underline{0.559}} & \textcolor{olive}{\textit{0.223}} \\
            CustomDiffusion~\cite{customdiffusion} & \textcolor{olive}{\textit{0.369}} & \textcolor{red}{\textbf{0.320}} & \textcolor{olive}{\textit{0.505}} & 0.169 \\
            FreeCustom~\cite{ding2024freecustom} & 0.360 & 0.289 & 0.480 & 0.142 \\
            OMG~\cite{kong2024omg} & 0.357 & 0.292 & 0.480 &  \textcolor{blue}{\underline{0.234}} \\
            \midrule
            \rowcolor[HTML]{FFFFE0} \textbf{ShowFlow-M} & \textcolor{red}{\textbf{0.454}} & \textcolor{blue}{\underline{0.314}} & \textcolor{red}{\textbf{0.575}} & \textcolor{red}{\textbf{0.306}} \\
            \bottomrule
    \end{tabular}
\end{subtable}

\end{table*}

\begin{figure*}[t!]
    \centering
    \includegraphics[width=0.95\linewidth]{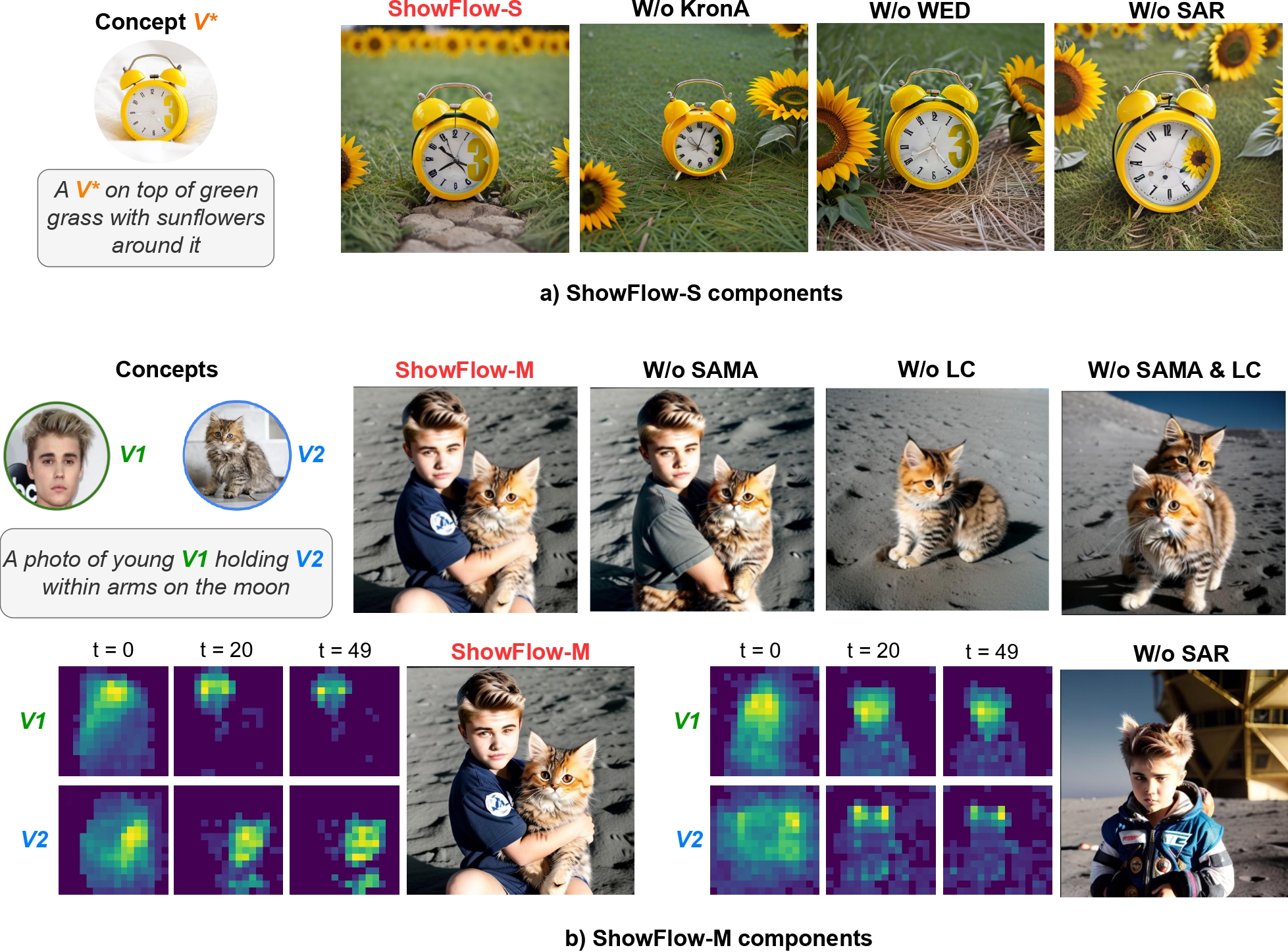}
    \caption{The effectiveness of components in ShowFlow.
    (a) ShowFlow-S: Kronecker Adapter (KronA), weight decomposition (WED) and Semantic-Aware Attention Regularization (SAR). (b) ShowFlow-M: SAMA, Layout Consistency (LC), and SAR objective in ShowFlow-S.
    }
    \label{figure:iclr_ablation-study-merge}

    
\end{figure*}

We present the results of the quantitative evaluation for the generation of a single concept in Table~\ref{table:chapter6_exps_single_quant}. Our proposed ShowFlow-S component exhibits comparable results in both DINO and CLIP-T metrics, thereby demonstrating its capability in balancing the trade-off of reconstruction and editability. ShowFlow-S also surpasses the baseline methods in ArcFace~\cite{deng2019arcface} score, further demonstrating its effectiveness in preserving human facial details.
The results of qualitative comparisons are shown in Figure~\ref{figure:chapter6_single_qualitative}. 




For multiple concepts generation, quantitative evaluation results are depicted in Table~\ref{table:chapter6_multi_quan_compare}. We showcase qualitative comparisons in Figure~\ref{figure:chapter6_multi_qual_baseline}. ShowFlow-M demonstrates superior performance in identity preservation compared to other methods, as measured by both DINO and ArcFace scores.
Regarding prompt alignment, Custom Diffusion~\cite{customdiffusion} scores highest as it generates broad-view images, while our method focuses on close-up views, leading to slightly lower scores. 

\section{Ablation Study}

\subsection{ShowFlow-S}

\vspace{0.2cm}
\noindent
\textbf{Effectiveness of components.} We consistently apply the disentangled learning~\cite{chendisenbooth} and evaluate the effectiveness of our proposed KronA-WED adapter and Semantic-Aware Attention Regularization (SAR) training objective. KronA-WED is evaluated by analyzing its Kronecker Adapter (KronA)~\cite{krona} and weight decomposition (WED) components.
As shown in Table~\ref{table:chapter6_exps_single_ablation}, by substituting the LoRA~\cite{lora} adapter with the KronA~\cite{krona} adapter, we significantly enhance the reconstruction capability of ShowFlow-S.
Moreover, WED enhances the learning capability of the model, thereby boosting both the DINO and CLIP-T scores. Together, these results highlight the advantage of the KronA-WED adapter. Notably, the proposed SAR objective improves identity preservation while slightly decreasing the prompt alignment score, and the decrease mainly occurs in prompts involving style or property changes. By decoupling the attention maps of noun and adjective tokens, SAR enables better alignment on interaction prompts, which is crucial for ShowFlow-M. Figure~\ref{figure:iclr_ablation-study-merge}a showcases illustrations for these evaluations.

\noindent
\begin{table*}[t!]
\centering
\caption{Ablation study for the effectiveness of components in ShowFlow-S and ShowFlow-M.}

\begin{subtable}[t]{.48\linewidth}
    \centering
    \small
    \caption{\textbf{ShowFlow-S:} Kronecker Adapter (KronA), weight decomposition (WED), and SAR training objective.}
    \label{table:chapter6_exps_single_ablation}
    \begin{tabular}{lcccc}
    \toprule
         & \textbf{DINO $\uparrow$} & \textbf{CLIP-T $\uparrow$} \\ 
        \hline
        \textbf{ShowFlow-S}       & \textbf{0.682} & 0.282 \\ 
        \hline
        W/o KronA & 0.647 \textcolor{red}{(-0.035)} & 0.283 \textcolor{blue}{(+0.001)} \\
        W/o WED  & 0.668 \textcolor{red}{(-0.014)} & 0.278 \textcolor{red}{(-0.004)} \\
        W/o SAR    & 0.660 \textcolor{red}{(-0.022)} & \textbf{0.284} \textcolor{blue}{(+0.002)} \\
        \bottomrule
    \end{tabular}
    
    \vspace{0.2cm}
    
    \begin{tabular}{lcccc}
    \toprule
         \textbf{Prompt Types }& \textbf{CLIP-T $\uparrow$} & \textbf{CLIP-T W/o SAR $\uparrow$} \\ 
        \hline
        Recontextualization & \textbf{0.756}           & 0.752 \textcolor{red}{(-0.004)}              \\ 
Restylization       & 0.678                    & \textbf{0.708} \textcolor{blue}{(+0.03)}      \\ 
Interaction         & \textbf{0.711}           & 0.701 \textcolor{red}{(-0.01)}               \\ 
Property change    & 0.642                    & \textbf{0.658} \textcolor{blue}{(+0.016)}     \\
        \bottomrule
    \end{tabular}
\end{subtable}%
\hfill
\begin{subtable}[t]{.48\linewidth}
    \centering
    \small
    \caption{\textbf{ShowFlow-M:} SAMA, Layout Consistency (LC) guidance, and SAR training object in ShowFlow-S.}
    \label{table:chapter6_multi_ablation}
    \begin{tabular}{lcccc}
        \toprule
         & \textbf{DINO $\uparrow$} & \textbf{CLIP-T $\uparrow$} \\ 
        \hline
        \textbf{ShowFlow-M}       & \textbf{0.454} & \textbf{0.314} \\ 
        \hline
        W/o SAMA \& LC & 0.435 \textcolor{red}{(-0.019)} & 0.311 \textcolor{red}{(-0.003)} \\
        W/o SAMA  & 0.431 \textcolor{red}{(-0.023)} & 0.312 \textcolor{red}{(-0.002)} \\
        W/o LC    & 0.442 \textcolor{red}{(-0.012)} & 0.31 \textcolor{red}{(-0.003)} \\
        \midrule
        W/o SAR & 0.440 \textcolor{red}{(-0.014)} & 0.307 \textcolor{red}{(-0.007)} \\
        \bottomrule
    \end{tabular}
\end{subtable}

\end{table*}

\begin{figure}[t!]
    \centering
    \includegraphics[width=\linewidth]{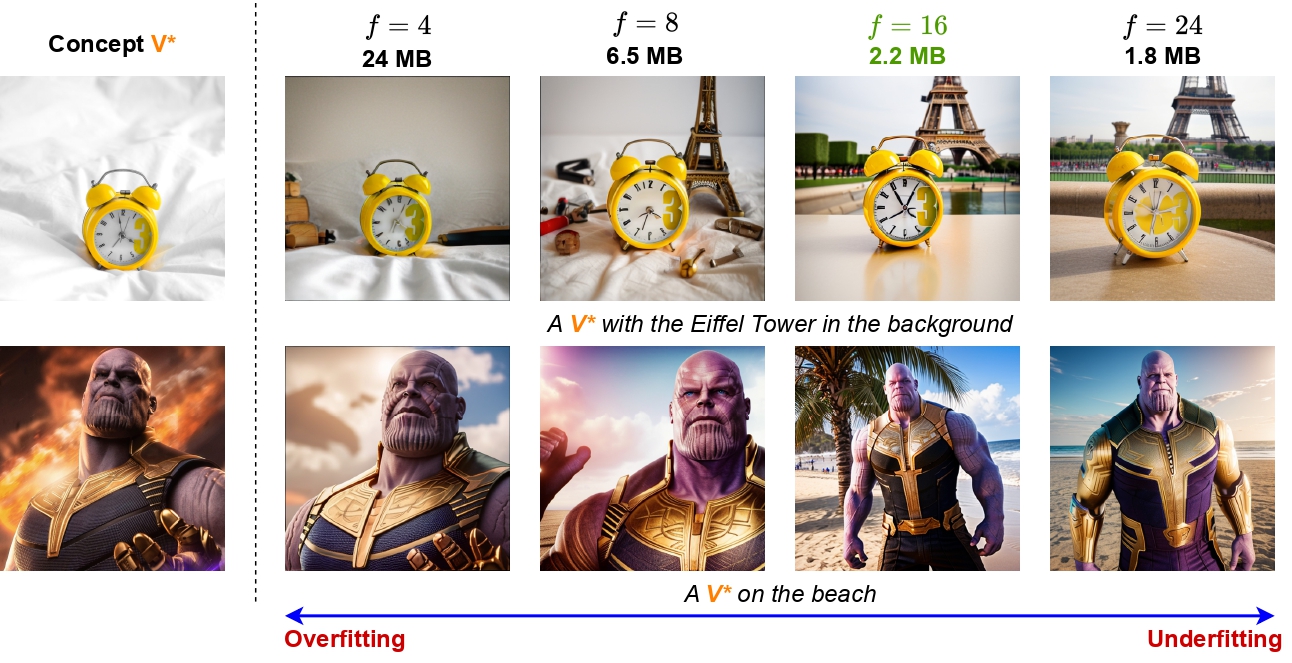}
     \caption{\revise{The effect of decomposition factor $f$ in KronA~\cite{krona} adapter for the concept learning and generation outputs. We provide the model size below each factor value.}}
    \label{figure:chapter6_single_ablation_krona_factor}
\end{figure}

\begin{figure}[t!]
    \centering
    \includegraphics[width=\linewidth]{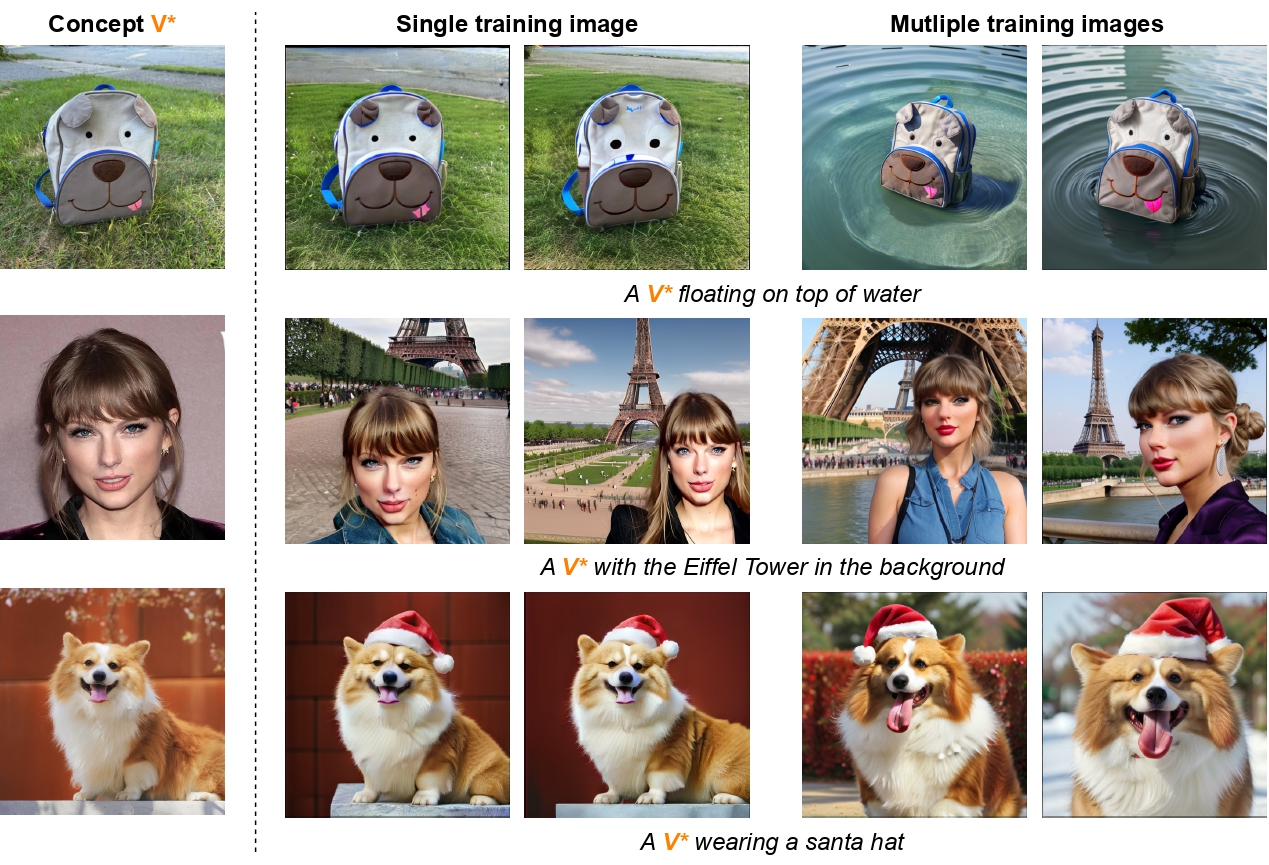}
     \caption{\revise{Generation of single concept using ShowFlow-S in different number of training images. For multiple images, we use from 3-5 images in our experiments.}}
    \label{figure:chapter6_single_ablation_num_train_imgs}
\end{figure}

\vspace{0.2cm}
\noindent
\textbf{Kronecker decomposition factor.}
Despite the fact that the rank of updated matrices obtained from KronA~\cite{krona} adapter is not depend on the factorization matrices, choosing the right value for decomposition values is an important detail.
The minimum number of paramerers in a KronA adapter is as follow:
\begin{equation}
    a_1a_2 + b_1b_2 = a_1a_2 + \frac{d}{a_1} \frac{k}{a_2} \geq 2 \sqrt{dk},
\end{equation}
and the equation holds when $a_1 = \sqrt{d}$ and $a_2 = \sqrt{k}$. 
\revise{Instead of choosing value for both $a_1$ and $a_2$, we follow Yeh~\etal~\cite{lycoris} to reduce them to a single decomposition factor $f$, and the value of $a_1$, $b_1$ will be specified based on $d$ as follow (similar to $a_2$, $b_2$ and $k$):
\begin{equation}
    a_1 = \max (u \leq \min(f, \sqrt{d}) \;|\; d \text{ mod } u = 0), \quad b_1 = \frac{d}{a_1}.
    \label{equation:chapter6_preliminary_krona_factor}
\end{equation}
}

\revise{Here, increasing the value of $f$ results in fewer parameters and vice versa. This variation may cause issues of underfitting (insufficient to capture the fine-grained details) or overfitting to the training images, which are illustrated in Figure~\ref{figure:chapter6_single_ablation_krona_factor}. In all of our experiments, we set the value $f$ to 16 to balance the efficiency and the effectiveness of ShowFlow-S.}

\begin{figure*}[t!]
	\centering
	\includegraphics[width=0.75\linewidth]{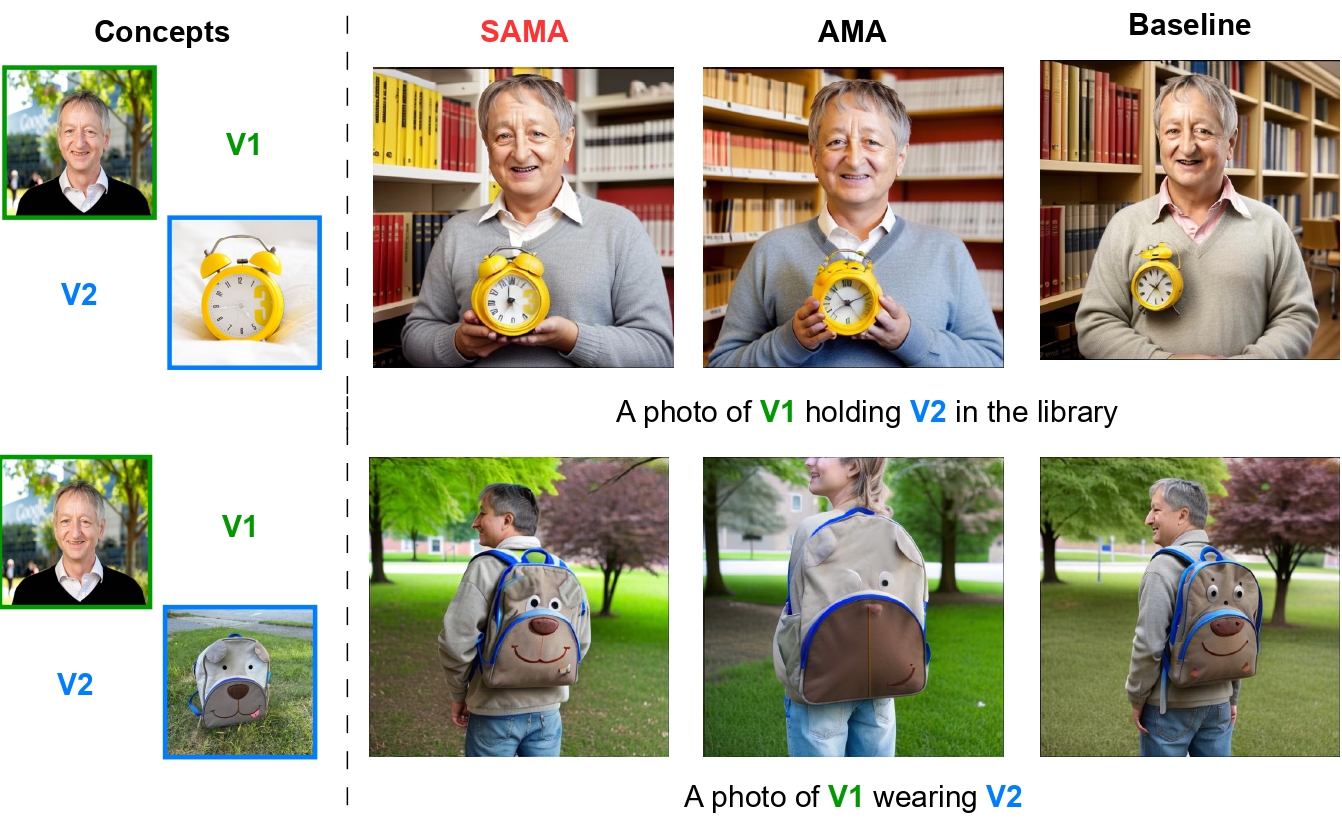}
	\caption{Comparison of the effectiveness of our proposed SAMA module in preserving intricate concept details versus the AMA module in DreamMatcher~\cite{nam2024dreammatcher}.}
	\label{appendix_figure:effectiveness_SAMA}
\end{figure*}

\vspace{0.1cm}
\noindent
\revise{\textbf{Number of training images.} 
We present the generation results of ShowFlow-S in scenarios involving both a single training image and multiple training images (ranging from 3 to 5 images used in our experiments) in Figure~\ref{figure:chapter6_single_ablation_num_train_imgs}.  
For concepts that do not contain many complex details such as regular humans and animals, the results indicate that ShowFlow-S can generate desirable images with just one training image. 
However, in this setting, the model tends to overfit the training image to preserve the concept's identity. Moreover, the diversity of the outcomes may decrease due to overfitting to the concept layout.
}




\subsection{ShowFlow-M}



\vspace{0.2cm}
\noindent
\textbf{Effectiveness of components.} Table~\ref{table:chapter6_multi_ablation} and Figure~\ref{figure:iclr_ablation-study-merge}b present the experiments results to demonstrate the effectiveness of SAMA and Layout Consistentcy guidance.
Applying SAMA significantly enhances the identity of each concept. However, without Layout Consistency guidance, one concept is often missed, leading to the issue of semantic matching and a decline in both metrics.
Moreover, simply sampling from fused weights can lead to significant identity degradation and missing concepts. 

\begin{table}[t!]
    \caption{Comparison between SAMA in ShowFlow-S and AMA of DreamMatcher~\cite{nam2024dreammatcher}. \textbf{Bold} values indicate the top score.}
    \label{appendix_table:rebuttal_comparison_sama_and_ama}
    \centering
    \renewcommand{\arraystretch}{1.1}
    \setlength{\tabcolsep}{6pt}
    \begin{tabular}{lccc}
        \toprule
        & \textbf{DINO $\uparrow$} & \textbf{CLIP-T $\uparrow$} & \textbf{ArcFace $\uparrow$} \\ 
        \midrule
        AMA~\cite{nam2024dreammatcher} & 0.442 & 0.728 & 0.236 \\
        \rowcolor[HTML]{FFFFE0} \textbf{SAMA} & \textbf{0.454} & \textbf{0.784} & \textbf{0.306} \\
        \bottomrule
    \end{tabular}
\end{table}



\vspace{0.2cm}
\noindent
\textbf{Utilizing ShowFlow-S with SAR objective for single concept learning.} 
In Figure~\ref{figure:iclr_ablation-study-merge}b, we display the generated images and the attention map of concept tokens when applying ShowFlow-M to models learned through ShowFlow-S, both with and without SAR. 
Without SAR, the maps for both concept tokens are initially unfocused during the early denoising steps, thereby causing our Layout Consistency guidance to struggle with maintaining the layout in later steps.
Consequently, this leads to vague masks for semantic matching, causing a decline in identity preservation and prompt alignment, as indicated in Table~\ref{table:chapter6_multi_ablation}. This observation highlights the importance of SAR in ShowFlow-S for enabling effective multi-concept generation in ShowFlow-M, demonstrating that ShowFlow forms a coherent and well-structured framework.

\vspace{0.2cm}
\noindent
\textbf{Comparisons between SAMA in ShowFlow-S and AMA in DreamMatcher. }
\label{rebuttal_section:sama_vs_ama}
To ensure accurate feature matching between reference and target semantic features, SAMA masks the subject’s features within the target feature to compute the matching cost volume \(\mathbf{C}\) more precisely. This refinement enhances correspondence estimation. To focus on semantics and structure, SAMA is applied selectively to layers containing semantic appearance information. 

In the qualitative results shown in Figure~\ref{appendix_figure:effectiveness_SAMA}, the generated image using SAMA for the prompt \textit{"A photo of \textcolor{olive}{V1} holding \textcolor{blue}{V2} in a library"} successfully preserves the detailed "3" on the clock face, whereas others fail. Similarly, for the prompt \textit{"A photo of \textcolor{olive}{V1} wearing \textcolor{blue}{V2}"}, details of the backpack, such as the ears and mouth, are completely lost without SAMA and remain unrecognizable in AMA. Moreover, as shown in Table~\ref{appendix_table:rebuttal_comparison_sama_and_ama}, SAMA significantly outperforms AMA~\cite{nam2024dreammatcher} in preserving identity.

\begin{figure*}[t!]
	\centering
	\includegraphics[width=\linewidth]{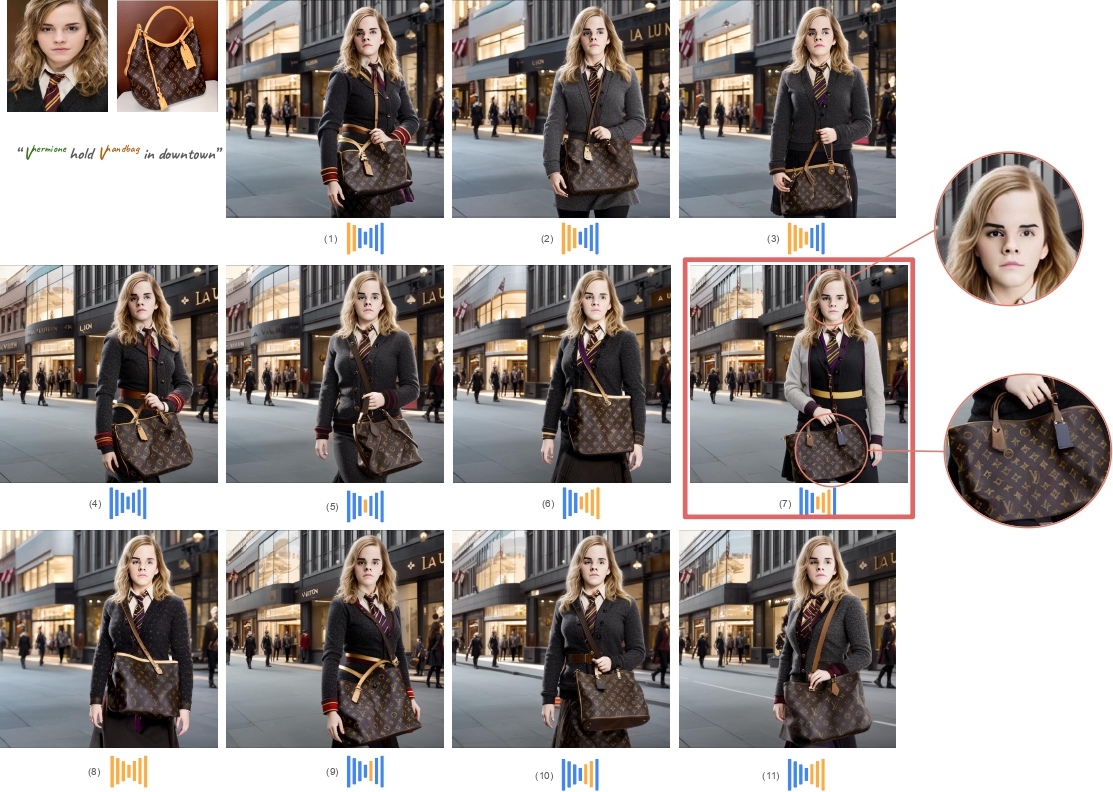}
	\caption{\revise{The impact of the SAMA module when implemented across various blocks in a U-Net architecture. The blue color signifies the original basic block, while the yellow color denotes the basic block in which self-attention is substituted by SAMA.}}
	\label{fig:chapter6_select_sama}
\end{figure*}



\vspace{0.2cm}
\noindent
\revise{
\textbf{Selective applying SAMA module in ShowFlow-M. }
\label{appendix:sama}
Experimental results demonstrate that substituting all self-attention modules with the SAMA module does not yield satisfactory results due to the layer-specific sensitivity of features. Since the estimated correspondence \(\mathbf{F}_k^{ref \rightarrow trg}\) is calculated from the cost volume \(\mathbf{C}_k\) derived from early decoder layers, which focus on semantics and structures~\cite{tumanyan2023plug,zhang2024tale,mou2024dragondiffusion}, SAMA should be applied to layers where spatial features have significant semantic appearance information. Through empirical observation, we achieve superior results when applying SAMA to the layers of the middle block and earlier blocks of the decoder in U-Net~\cite{unet} instead of replacing all self-attention modules in the 7 basic blocks with SAMA, as shown in Figure  
\ref{fig:chapter6_select_sama}. }



\revise{
\section{Computational Efficiency of ShowFlow}

We evaluate the computational efficiency of ShowFlow for both single- and multi-concept generation. For single-concept settings, we report the training time and memory required to obtain adapter weights across all methods. For multi-concept settings, we consider the adapter fusion cost, generation time, and corresponding memory usage. Detailed comparisons on a single NVIDIA A100 40GB GPU are provided as follow:

\vspace{0.2cm}
\noindent
\textbf{Single-concept learning.}
It is noteworthy that ShowFlow-S adopts an image adapter from DisenBooth~\cite{chendisenbooth} and additionally stores attention maps of concept tokens for the proposed SAR objective. The training time and GPU memory usage are reported in Table~\ref{tab:response_mem_single}. Compared to prior parameter-efficient methods, ShowFlow-S introduces a moderate increase in memory usage due to the additional adapter components and attention regularization. However, it remains significantly more efficient than full fine-tuning approaches (\eg, DreamBooth), while achieving stronger identity preservation and editability.

\begin{table*}[h]
\centering
\caption{\revise{Training time and memory requirement of ShowFlow-S compared with baselines.}}
\label{tab:response_mem_single}
\revise{
\begin{tabular}{lccp{7.5cm}}
\toprule
\textbf{Method}            & \textbf{Training Time} & \textbf{Training Memory} & \textbf{Note}                   \\
\midrule
DreamBooth~\cite{ruiz2023dreambooth}        & 10 min        & 21 GB           & Fine-tune the full SD U-Net                                     \\
Custom Diffusion~\cite{customdiffusion}  & 6 min         & 16.5 GB         & Fine-tune K-V in cross-attention layers                      \\
DisenBooth~\cite{chendisenbooth}        & 15 min        & 8.8 GB          & LoRA ($r=4$) + image adapter                 \\
ED-LoRA~\cite{gu2024mix}           & 8 min         & 8.5 GB          & LoRA ($r=4$) \\
LoKr~\cite{lycoris}              & 6 min         & 6 GB            & KronA ($f=16$)                        \\
\midrule
\rowcolor[HTML]{FFFFE0} \textbf{ShowFlow-S} & 15 min        & 10 GB           & KronA-WED ($f=16$) + image adapter + SAR\\
\bottomrule
\end{tabular}}
\end{table*}

\vspace{0.2cm}
\noindent
\textbf{Multi-concept generation.}
To generate an image for a prompt involving $N$ concepts, ShowFlow-M first adopts Gradient Fusion~\cite{gu2024mix} to merge weights from $N$ adapters (\textit{offline}). Due to the need for reference features, it has to produce $(N+1)$ images (one target image and $N$ reference images), introducing additional memory and computational overhead in the initial pass. However, the resulting reference features can be \textit{reused} across prompts for the same concepts. As a result, subsequent runs are significantly faster, making the approach suitable for large-scale, real-world applications. The fusion cost and efficiency for $N=2$ are summarized in Table~\ref{tab:response_mem_multi}.

\begin{table*}[t!]
\centering
\caption{\revise{Fusion and generation efficiency of ShowFlow-M compared with baselines in a two-concept setting. Values for ShowFlow-M are reported as initial and subsequent runs.}}
\label{tab:response_mem_multi}
\revise{
\begin{tabular}{lcccc}
\toprule
\textbf{Method}            & \begin{tabular}[c]{@{}c@{}}\textbf{Fusion Time}\\ \textit{(offline)}\end{tabular} & \begin{tabular}[c]{@{}c@{}}\textbf{Fusion Memory}\\ \textit{(offline)}\end{tabular} & \begin{tabular}[c]{@{}c@{}}\textbf{Generation Time}\\ \textit{(online)}\end{tabular} & \begin{tabular}[c]{@{}c@{}}\textbf{Generation Memory}\\ \textit{(online)}\end{tabular} \\
\midrule
Mix-Of-Show~\cite{gu2024mix}       & 5 min       & 16.4 GB       & 2 s           & 6.4 GB          \\
CustomDiffusion~\cite{customdiffusion}   & 2 min       & 0             & 1.75 s        & 6.2 GB          \\
FreeCustom~\cite{ding2024freecustom}        & 0           & 0             & 20 s          & 19 GB           \\
OMG~\cite{kong2024omg}               & 0           & 0             & 15 s          & 13.5 GB         \\
\midrule
\rowcolor[HTML]{FFFFE0} \textbf{ShowFlow-M} & 5 min       & 16.4 GB       & 9.3 s         & 25.7 GB  \\ 
\rowcolor[HTML]{FFFFE0} \textbf{ShowFlow-M (Optimized)} & 5 min       & 16.4 GB       & 2.75 s         & 10 GB  \\ 
\bottomrule
\end{tabular}}
\end{table*}


\vspace{0.1cm}
Overall, ShowFlow achieves a favorable balance between efficiency and performance: it remains substantially more efficient than full fine-tuning approaches, while introducing a controlled overhead to support more robust and accurate personalized generation. ShowFlow-M is particularly well suited for real-world, large-scale generation scenarios. In future work, we aim to improve efficiency by reducing redundant computations in multi-concept sampling and designing a more lightweight reference features injection strategy, enabling faster and more memory-efficient deployment without compromising compositional fidelity.
}

\section{User Study}
\label{section:chapter6_single_user-study}

We invited 20 participants (16 males and 4 females) with age range from 18 to 25 from our research community and university to participate in the study. Half of them have a background in AI, and some are acquainted with our evaluation metrics. They brought diverse perspectives to the evaluation process, ensuring an objective assessment. 

We collected 1000 responses from participants, where each response on a scale from 1 (very bad) to 5 (very good). To ensure fair and consistent assessments, we provided a reference document that describes the criteria for each score. We also blinded the method names and shuffled the method outputs in each question.
Following previous studies~\cite{ruiz2023dreambooth, customdiffusion, chendisenbooth}, for single concept generation, we considered the metrics of \textit{identity preservation} and \textit{prompt alignment}. 
Regarding to multiple concepts generation, as the experimental evaluation metrics alone are not sufficiently expressive, we introduced a metric called \textit{naturalness of interaction} to measure how good the naturalness of interaction between the human and object (or animal) in the image
is, such as human pose, the size and the position of objects. 

\begin{figure*}[t!]
    \centering
    \begin{minipage}[b]{0.8\linewidth}
        \centering
        \includegraphics[width=\linewidth]{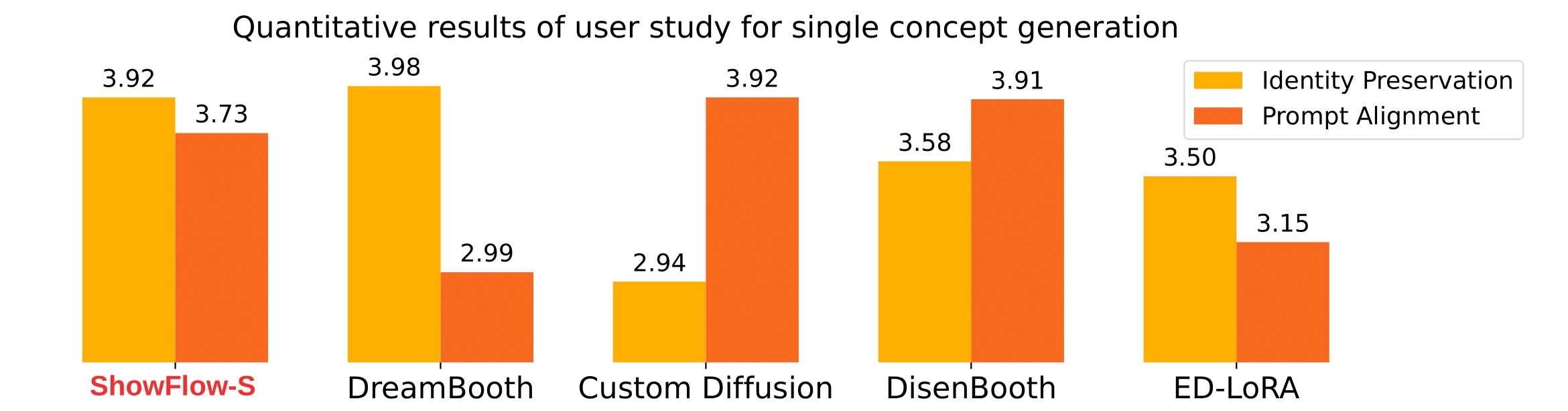}
        \caption{User study results: ShowFlow-S and baselines.}
        \label{fig:chapter7_userstudy_conceptflow_s}
    \end{minipage} \\ 
    \begin{minipage}[b]{0.8\linewidth}
        \centering
        \includegraphics[width=\linewidth]{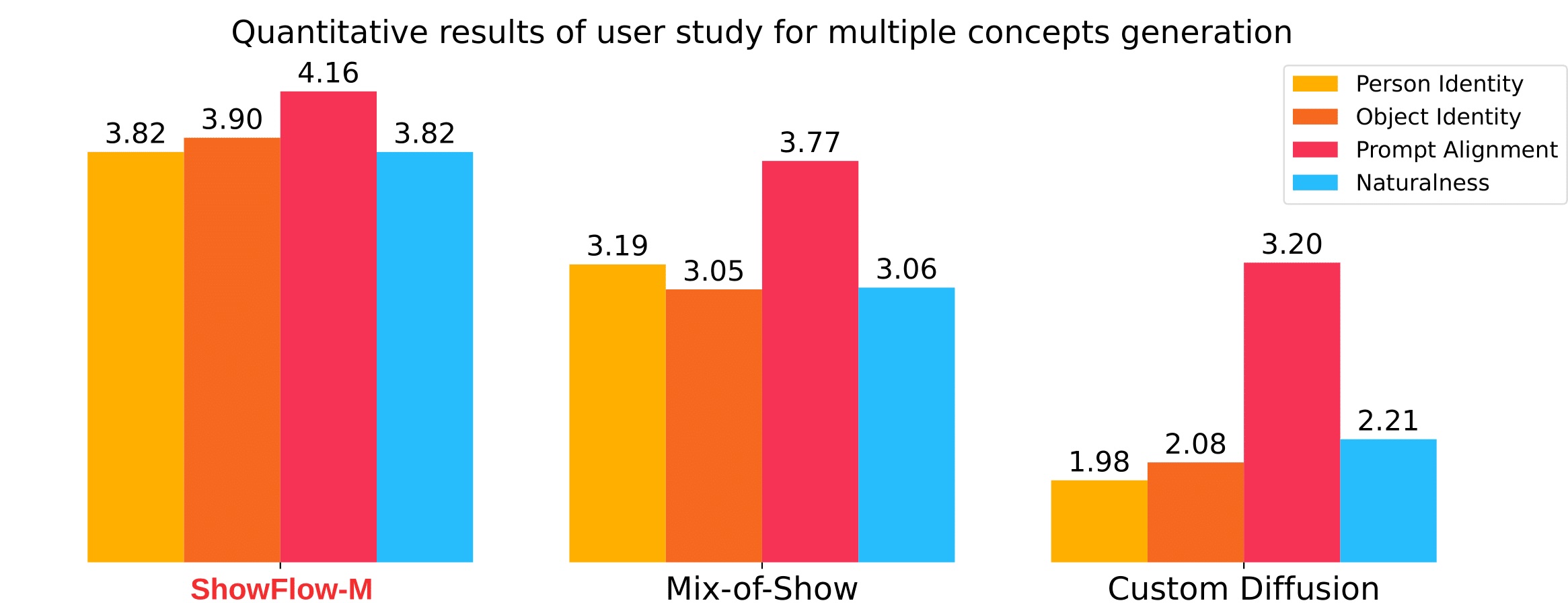}
        \caption{User study results: ShowFlow-M and baselines.}
        \label{fig:chapter7_userstudy_conceptflow_m}
    \end{minipage}
\end{figure*}

The results of our study on single concept generation are shown in Figure~\ref{fig:chapter7_userstudy_conceptflow_s}. They indicate that users were satisfied with ShowFlow-S in terms of both identity preservation (\ie, reconstruction) and prompt alignment (\ie, editability), with average scores of 3.92 and 3.73. 
For multiple concept generation in Figure~\ref{fig:chapter7_userstudy_conceptflow_m}, ShowFlow-M outperformed other methods across all metrics by significant margins.
Compared to the experiment quantitative results presented in Table~\ref{table:chapter6_multi_quan_compare}, the user study provides deeper insight into the methods' performance in generating multiple concepts.

\section{Potential Applications}
\begin{figure*}[t!]
    \centering
    \includegraphics[width=0.9\linewidth]{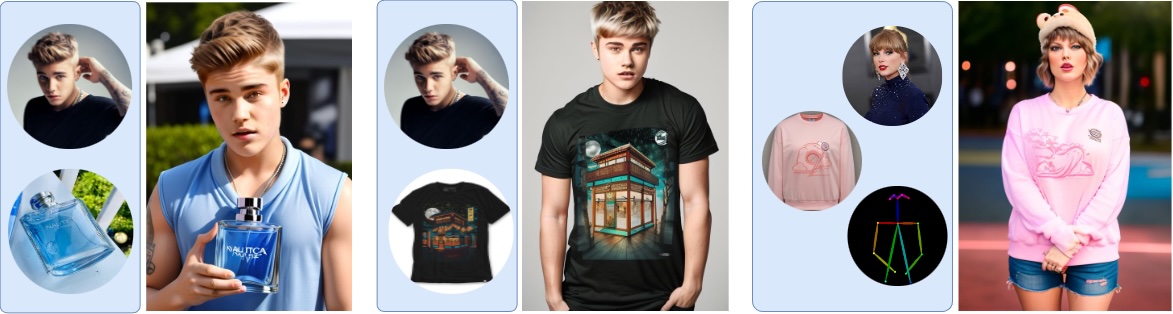}
     \caption{Illustration of ShowFlow on the virtual dressing and advertisement applications.}
    \label{figure:cvpr_application}
\end{figure*}

The high-fidelity and condition-free composition capabilities of ShowFlow naturally extend to practical real-world applications, as illustrated in Figure~\ref{figure:cvpr_application}. For \textbf{virtual try-on}, ShowFlow can seamlessly synthesize specific garments onto diverse human subjects and poses, facilitating realistic fitting experiences without requiring manual spatial masks. Furthermore, in the domain of \textbf{product advertisement}, ShowFlow robustly preserves identity details within complex compositional layouts, enabling the automated generation of tailored, high-quality promotional content across diverse visual contexts.

\revise{
\section{Limitations and Discussion}
\label{appendix:limitation}

In this section, we analyze the limitations of ShowFlow from two perspectives: (i) single-concept property and style modification in ShowFlow-S, and (ii) the compositional behavior of ShowFlow-M.

\subsection{Property and Style Changes in ShowFlow-S}

The primary limitation of ShowFlow-S lies in prompts involving \textit{property changes} (e.g., color modification) and \textit{restylization} (e.g., artistic transformations). As shown in Figure~\ref{fig:showflows_failedcase_analyze}, although the proposed KronA-WED and SAR improve identity preservation, SAR can suppress the attention of style or attribute tokens. Specifically, the attention maps of style tokens are often overshadowed by those of concept tokens ($V_{\text{rand}}$ and $V_{\text{class}}$) in regions corresponding to the subject, leading to weak or missing attribute modifications. Despite this limitation, ShowFlow-S demonstrates strong performance in interaction and recontextualization prompts (see Table~\ref{table:chapter6_exps_single_ablation}), which provides a solid foundation for multi-concept composition in ShowFlow-M.

\begin{figure*}[t!]
    \centering
    \includegraphics[width=\linewidth]{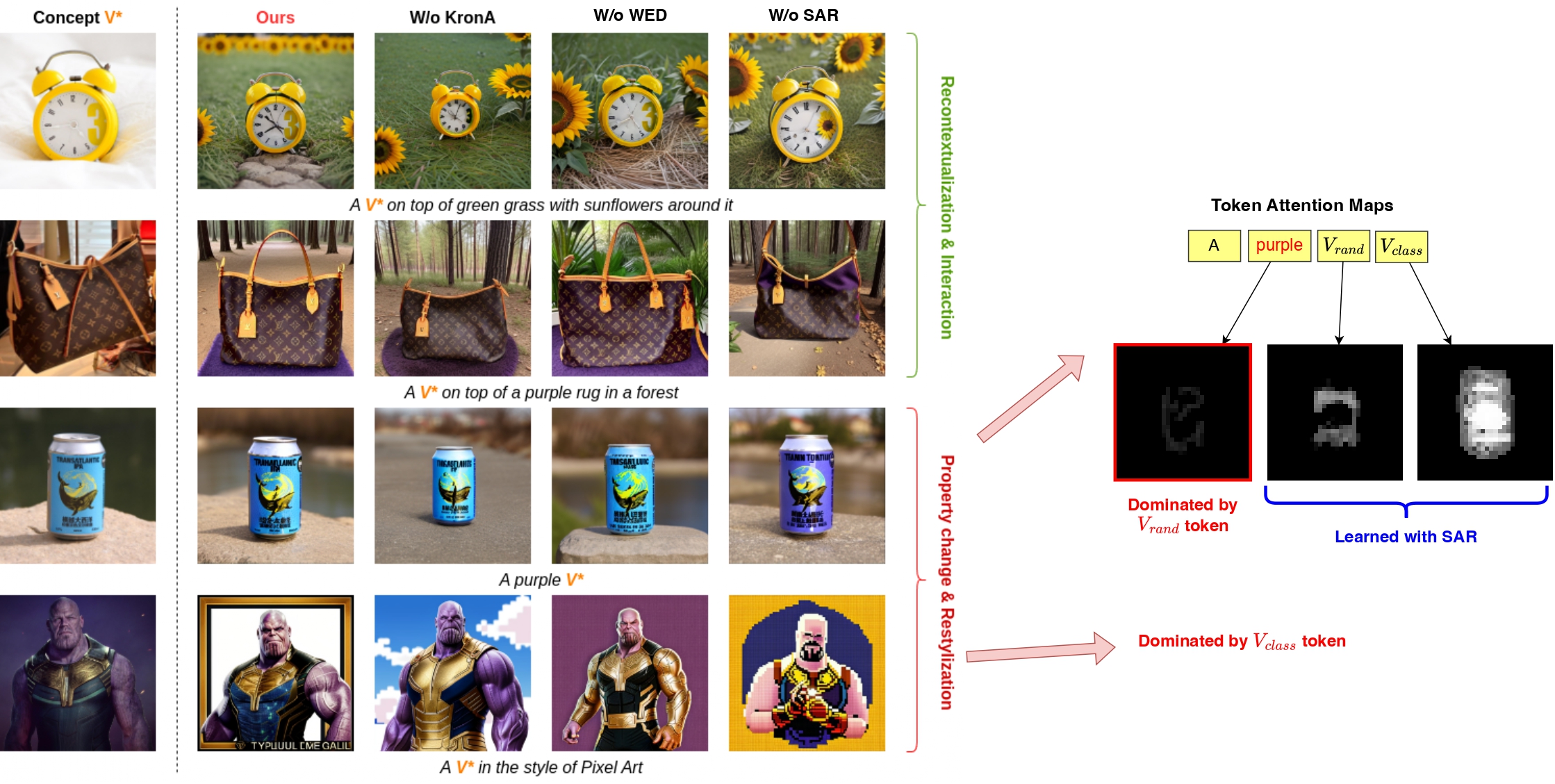}
    \caption{\revise{Failure cases of ShowFlow-S. While KronA-WED and SAR improve identity preservation, SAR may suppress style or attribute tokens, resulting in limited effectiveness for property modification and restylization tasks.}}
    \label{fig:showflows_failedcase_analyze}
\end{figure*}

\subsection{Challenging Cases for ShowFlow-M}

\vspace{0.2cm}
\noindent
\textbf{Semantically similar concepts generation.}
ShowFlow-M has limited capacity in the composition of semantically similar concepts (e.g., two humans, or visually similar animals such as a corgi and a husky). In contrast, heterogeneous combinations such as human–object and human–animal are handled more reliably.

This issue is due to the well-known \textit{attribute binding problem}~\cite{fengtraining, lee2023aligning} in Stable Diffusion 1.5~\cite{rombach2022high} and CLIP~\cite{clip}. As illustrated in Figure~\ref{fig:rebuttal_sematic_similar_analyze}, when concepts share similar semantic embeddings, the model often produces incorrect layouts early in the denoising process. Consequently, our Layout Consistency guidance becomes less effective, as it relies on early-stage attention maps to refine later steps.

\begin{figure*}[t!]
    \centering
    \includegraphics[width=\linewidth]{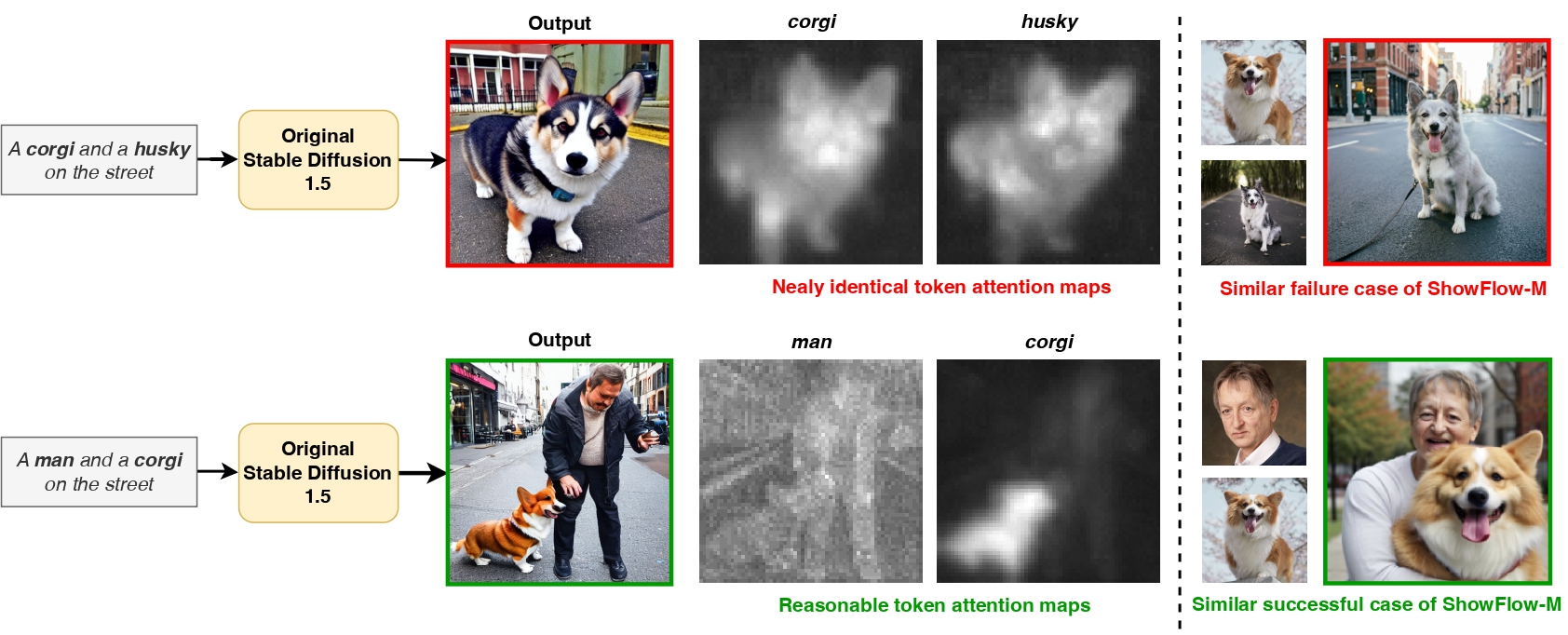}
    \caption{\revise{Failure cases of ShowFlow-M with semantically similar concepts. Due to attribute binding issues, the model produces incorrect layouts early in the denoising process, limiting the effectiveness of Layout Consistency guidance.}}
    \label{fig:rebuttal_sematic_similar_analyze}
\end{figure*}

\vspace{0.2cm}
\noindent
\textbf{Scaling to more than two concepts.}
As illustrated in Figure~\ref{fig:more_than_2_concepts}, increasing the number of concepts compromises the stability of ShowFlow-M. We attribute this to the architectural biases of Stable Diffusion 1.5~\cite{rombach2022high} and CLIP~\cite{clip}, which struggle with the multi-entity "counting" categories identified in the GenEval benchmark~\cite{ghosh2023geneval}. Our analysis of cross-attention maps (Figure~\ref{fig:sd_complex_prompt}) confirms that the base model experiences significant semantic entanglement in these scenarios.

\begin{figure*}[t!]
    \centering
    \includegraphics[width=0.9\linewidth]{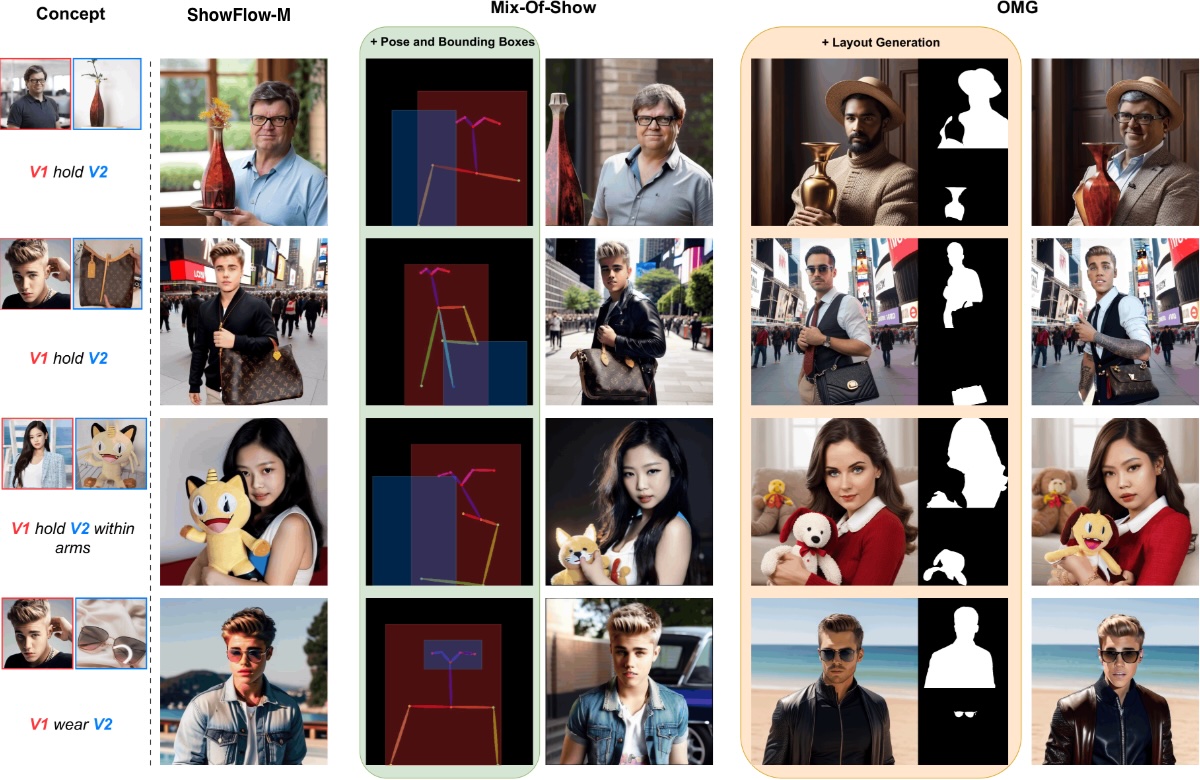}
    \caption{\revise{Comparison with condition-based methods. While condition-based approaches rely on spatial inputs, ShowFlow-M enables more flexible and natural object interactions without additional conditions.}}
    \label{fig:placeholder}
\end{figure*}

\begin{figure}[t]
    \centering
    \begin{subfigure}{\linewidth}
        \centering
        \includegraphics[width=1.0\linewidth]{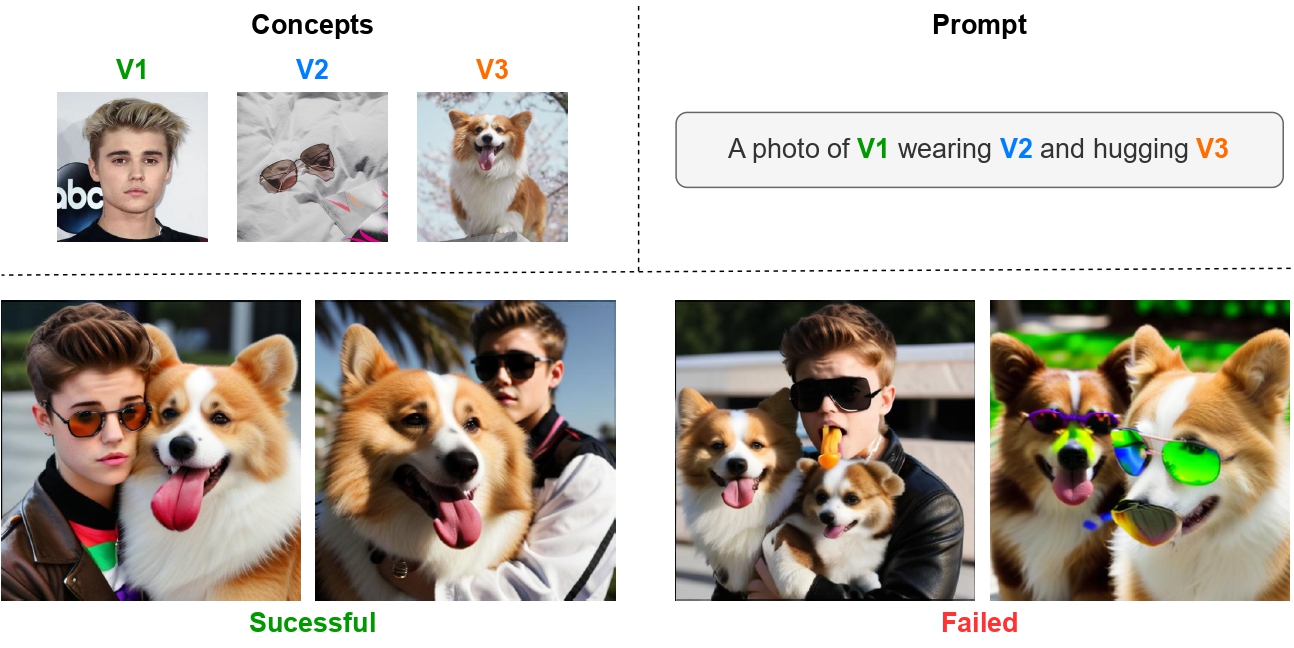}
        \caption{\revise{ShowFlow-M shows increasing inconsistency and identity degradation.}}
        \label{fig:more_than_2_concepts}
    \end{subfigure}
    
    \vspace{0.5cm} 
    
    \begin{subfigure}{\linewidth}
        \centering
        \includegraphics[width=1.0\linewidth]{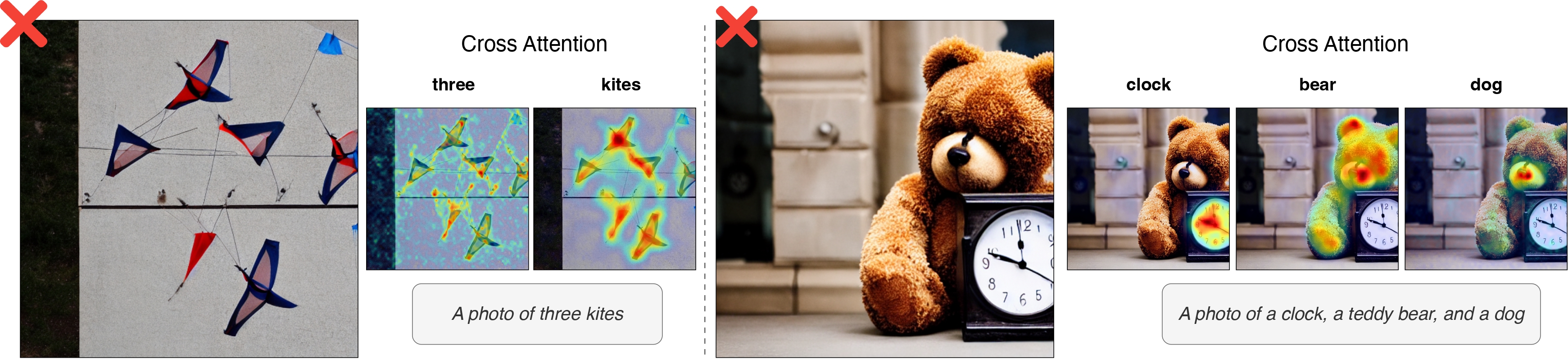}
        \caption{\revise{Failure cases of the base model on complex prompts.}}
        \label{fig:sd_complex_prompt}
    \end{subfigure}
    
    \caption{\revise{Analysis of failure cases of ShowFlow-M when handling complex multi-concept prompts.}}
    \label{fig:more_than_2_concepts_combined}
\end{figure}

\vspace{0.2cm}
\noindent
\textbf{Comparison with condition-based methods.}
Overall, the primary limitations of ShowFlow-M stem from the capacity of the underlying diffusion model. Prior works~\cite{gu2024mix, kong2024omg, ding2024freecustom} address these challenges by introducing additional spatial conditions (\eg, bounding boxes, pose skeletons, or segmentation masks). While effective, these approaches require extra inputs and often restrict the model's ability to generate \textit{natural object interactions}.

In contrast, ShowFlow-M operates in a condition-free manner and demonstrates strong performance in interaction-heavy scenarios such as human–animal and human–object compositions. As shown in Figure~\ref{fig:placeholder}, our method produces more natural and flexible interactions compared to condition-based approaches, highlighting its practical advantages despite the increased difficulty of the task.
}


\section{Conclusion}

In this paper, we present ShowFlow, a robust framework for the personalized image generation task. ShowFlow consists of two components: ShowFlow-S for single concept learning and generation, and ShowFlow-M for multi-concept generation. ShowFlow-S introduces the KronA-WED adapter and a novel Semantic-Aware Attention Regularization (SAR) training objective to balance the trade-off between reconstruction and editability. ShowFlow-M introduces SAMA and Layout Consistency guidance to combine concepts learned by ShowFlow-S, significantly enhancing the identity of each concept while addressing concept omissions without requiring additional conditions. We demonstrate the effectiveness of ShowFlow through extensive experiments and a user study. 

\section*{Acknowledgment}

This research is funded by Vietnam National Foundation for Science and Technology Development (NAFOSTED) under Grant Number 102.05-2023.31.

\printcredits
\bibliographystyle{cas-model2-names}
\balance
\bibliography{sn-bibliography}

@inproceedings{ruiz2023dreambooth,
  title={Dreambooth: Fine tuning text-to-image diffusion models for subject-driven generation},
  author={Ruiz, Nataniel and Li, Yuanzhen and Jampani, Varun and Pritch, Yael and Rubinstein, Michael and Aberman, Kfir},
  booktitle={CVPR},
  pages={22500--22510},
  year={2023}
}

@inproceedings{clip,
  title={Learning transferable visual models from natural language supervision},
  author={Radford, Alec and Kim, Jong Wook and Hallacy, Chris and Ramesh, Aditya and Goh, Gabriel and Agarwal, Sandhini and Sastry, Girish and Askell, Amanda  and others},
  booktitle={ICML},
  pages={8748--8763},
  year={2021},
  @organization={PMLR}
}

@inproceedings{rombach2022high,
  title={High-resolution image synthesis with latent diffusion models},
  author={Rombach, Robin and Blattmann, Andreas and Lorenz, Dominik and Esser, Patrick and Ommer, Bj{\"o}rn},
  booktitle={CVPR},
  pages={10684--10695},
  year={2022}
}

@inproceedings{dora,
  title={DoRA: Weight-Decomposed Low-Rank Adaptation},
  author={Liu, Shih-yang and Wang, Chien-Yi and Yin, Hongxu and Molchanov, Pavlo and Wang, Yu-Chiang Frank and Cheng, Kwang-Ting and Chen, Min-Hung},
  booktitle={ICML},
  year         = {2024}

}

@inproceedings{lora,
  title={LoRA: Low-Rank Adaptation of Large Language Models},
  author={Hu, Edward J and Wallis, Phillip and Allen-Zhu, Zeyuan and Li, Yuanzhi and Wang, Shean and Wang, Lu and Chen, Weizhu and others},
  booktitle={International Conference on Learning Representations},
  year         = {2024}
}

@InProceedings{nam2024dreammatcher,
    author    = {Nam, Jisu and Kim, Heesu and Lee, DongJae and Jin, Siyoon and Kim, Seungryong and Chang, Seunggyu},
    title     = {DreamMatcher: Appearance Matching Self-Attention for Semantically-Consistent Text-to-Image Personalization},
    booktitle = {CVPR},
    month     = {June},
    year      = {2024},
    pages     = {8100-8110}
}

@incollection{krona,
  title={KronA: Parameter-Efficient Tuning with Kronecker Adapter},
  author={Edalati, Ali and Tahaei, Marzieh and Kobyzev, Ivan and Nia, Vahid Partovi and Clark, James J and Rezagholizadeh, Mehdi},
  booktitle={Enhancing LLM Performance: Efficacy, Fine-Tuning, and Inference Techniques},
  pages={49--65},
  year={2025},
  publisher={Springer}
}

@article{gu2024mix,
  title={Mix-of-show: Decentralized low-rank adaptation for multi-concept customization of diffusion models},
  author={Gu, Yuchao and Wang, Xintao and Wu, Jay Zhangjie and Shi, Yujun and Chen, Yunpeng and Fan, Zihan and Xiao, Wuyou and Zhao, Rui and Chang, Shuning and Wu, Weijia and others},
  journal={NeurIPS},
  volume={36},
  year={2024}
}

@inproceedings{textualinversion,
  author       = {Rinon Gal and
                  Yuval Alaluf and
                  Yuval Atzmon and
                  Or Patashnik and
                  Amit Haim Bermano and
                  Gal Chechik and
                  Daniel Cohen{-}Or},
  title        = {An Image is Worth One Word: Personalizing Text-to-Image Generation
                  using Textual Inversion},
  booktitle    = {ICLR},
  year         = {2023},
}

@inproceedings{customdiffusion,
  title={Multi-concept customization of text-to-image diffusion},
  author={Kumari, Nupur and Zhang, Bingliang and Zhang, Richard and Shechtman, Eli and Zhu, Jun-Yan},
  booktitle={CVPR},
  pages={1931--1941},
  year={2023}
}

@inproceedings{chendisenbooth,
  title={DisenBooth: Identity-Preserving Disentangled Tuning for Subject-Driven Text-to-Image Generation},
  author={Chen, Hong and Zhang, Yipeng and Wu, Simin and Wang, Xin and Duan, Xuguang and Zhou, Yuwei and Zhu, Wenwu},
  booktitle={ICLR},
  year={2024}

}

@article{pplus,
  title={p+: Extended textual conditioning in text-to-image generation},
  author={Voynov, Andrey and Chu, Qinghao and Cohen-Or, Daniel and Aberman, Kfir},
  journal={arXiv preprint arXiv:2303.09522},
  year={2023}
}

@article{loracommunity,
  title={Lora: Low-rank adaptation of large language models.},
  author={Hu, Edward J and Shen, Yelong and Wallis, Phillip and Allen-Zhu, Zeyuan and Li, Yuanzhi and Wang, Shean and Wang, Lu and Chen, Weizhu and others},
  journal={ICLR},
  volume={1},
  number={2},
  pages={3},
  year={2022}
}

@inproceedings{lycoris,
  title={Navigating text-to-image customization: From lycoris fine-tuning to model evaluation},
  author={Yeh, Shih-Ying and Hsieh, Yu-Guan and Gao, Zhidong and Yang, Bernard BW and Oh, Giyeong and Gong, Yanmin},
  booktitle={ICLR},
  year={2023}
}

@inproceedings{caron2021emerging,
  title={Emerging properties in self-supervised vision transformers},
  author={Caron, Mathilde and Touvron, Hugo and Misra, Ishan and J{\'e}gou, Herv{\'e} and Mairal, Julien and Bojanowski, Piotr and Joulin, Armand},
  booktitle={ICCV},
  pages={9650--9660},
  year={2021}
}

@inproceedings{unet,
  title={U-net: Convolutional networks for biomedical image segmentation},
  author={Ronneberger, Olaf and Fischer, Philipp and Brox, Thomas},
  booktitle={Medical image computing and computer-assisted intervention--MICCAI 2015: 18th international conference, Munich, Germany, October 5-9, 2015, proceedings, part III 18},
  pages={234--241},
  year={2015},
  organization={Springer}
}

@inproceedings{tumanyan2023plug,
  title={Plug-and-play diffusion features for text-driven image-to-image translation},
  author={Tumanyan, Narek and Geyer, Michal and Bagon, Shai and Dekel, Tali},
  booktitle={CVPR},
  pages={1921--1930},
  year={2023}
}

@inproceedings{
mou2024dragondiffusion,
title={DragonDiffusion: Enabling Drag-style Manipulation on Diffusion Models},
author={Chong Mou and Xintao Wang and Jiechong Song and Ying Shan and Jian Zhang},
booktitle={The Twelfth International Conference on Learning Representations},
year={2024},
url={https://openreview.net/forum?id=OEL4FJMg1b}
}

@inproceedings{agarwal2023star,
  title={A-star: Test-time attention segregation and retention for text-to-image synthesis},
  author={Agarwal, Aishwarya and Karanam, Srikrishna and Joseph, KJ and Saxena, Apoorv and Goswami, Koustava and Srinivasan, Balaji Vasan},
  booktitle={Proceedings of the IEEE/CVF International Conference on Computer Vision},
  pages={2283--2293},
  year={2023}
}

@inproceedings{avrahami2023break,
  title={Break-a-scene: Extracting multiple concepts from a single image},
  author={Avrahami, Omri and Aberman, Kfir and Fried, Ohad and Cohen-Or, Daniel and Lischinski, Dani},
  booktitle={SIGGRAPH Asia},
  pages={1--12},
  year={2023}
}

@inproceedings{han2023svdiff,
  title={Svdiff: Compact parameter space for diffusion fine-tuning},
  author={Han, Ligong and Li, Yinxiao and Zhang, Han and Milanfar, Peyman and Metaxas, Dimitris and Yang, Feng},
  booktitle={ICCV},
  pages={7323--7334},
  year={2023}
}

@inproceedings{kong2024omg,
  title={Omg: Occlusion-friendly personalized multi-concept generation in diffusion models},
  author={Kong, Zhe and Zhang, Yong and Yang, Tianyu and Wang, Tao and Zhang, Kaihao and Wu, Bizhu and Chen, Guanying and Liu, Wei and Luo, Wenhan},
  booktitle={ECCV},
  pages={253--270},
  year={2025},
  organization={Springer}
}

@article{zhang2024tale,
  title={A tale of two features: Stable diffusion complements dino for zero-shot semantic correspondence},
  author={Zhang, Junyi and Herrmann, Charles and Hur, Junhwa and Polania Cabrera, Luisa and Jampani, Varun and Sun, Deqing and Yang, Ming-Hsuan},
  journal={Advances in Neural Information Processing Systems},
  volume={36},
  year={2024}
}

@inproceedings{he2015delving,
  title={Delving deep into rectifiers: Surpassing human-level performance on imagenet classification},
  author={He, Kaiming and Zhang, Xiangyu and Ren, Shaoqing and Sun, Jian},
  booktitle={ICCV},
  pages={1026--1034},
  year={2015}
}

@article{xiao2024fastcomposer,
  title={Fastcomposer: Tuning-free multi-subject image generation with localized attention},
  author={Xiao, Guangxuan and Yin, Tianwei and Freeman, William T and Durand, Fr{\'e}do and Han, Song},
  journal={International Journal of Computer Vision},
  pages={1--20},
  year={2024},
  publisher={Springer}
}

@inproceedings{deng2019arcface,
  title={Arcface: Additive angular margin loss for deep face recognition},
  author={Deng, Jiankang and Guo, Jia and Xue, Niannan and Zafeiriou, Stefanos},
  booktitle={CVPR},
  pages={4690--4699},
  year={2019}
}

@inproceedings{ding2024freecustom,
  title={FreeCustom: Tuning-Free Customized Image Generation for Multi-Concept Composition},
  author={Ding, Ganggui and Zhao, Canyu and Wang, Wen and Yang, Zhen and Liu, Zide and Chen, Hao and Shen, Chunhua},
  booktitle={CVPR},
  pages={9089--9098},
  year={2024}
}

@article{alaluf2023neural,
  title={A neural space-time representation for text-to-image personalization},
  author={Alaluf, Yuval and Richardson, Elad and Metzer, Gal and Cohen-Or, Daniel},
  journal={ACM Transactions on Graphics (TOG)},
  volume={42},
  number={6},
  pages={1--10},
  year={2023},
  publisher={ACM New York, NY, USA}
}

@inproceedings{po2024orthogonal,
  title={Orthogonal adaptation for modular customization of diffusion models},
  author={Po, Ryan and Yang, Guandao and Aberman, Kfir and Wetzstein, Gordon},
  booktitle={Proceedings of the IEEE/CVF conference on computer vision and pattern recognition},
  pages={7964--7973},
  year={2024}
}

@article{loracomposer,
  author={Yang, Yang and Wang, Wen and Peng, Liang and Song, Chaotian and Chen, Yao and Li, Hengjia and Yang, Xiaolong and Lu, Qinglin and Cai, Deng and He, Xiaofei and Wu, Boxi and Liu, Wei},
  journal={IEEE Transactions on Image Processing}, 
  title={LoRA-Composer: Leveraging Low-Rank Adaptation for Multi-Concept Customization in Training-Free Diffusion Models}, 
  year={2025},
  volume={34},
  number={},
  pages={8145-8158},
  doi={10.1109/TIP.2025.3633153}}

@article{chefer2023attend,
  title={Attend-and-excite: Attention-based semantic guidance for text-to-image diffusion models},
  author={Chefer, Hila and Alaluf, Yuval and Vinker, Yael and Wolf, Lior and Cohen-Or, Daniel},
  journal={ACM transactions on Graphics (TOG)},
  volume={42},
  number={4},
  pages={1--10},
  year={2023},
  publisher={ACM New York, NY, USA}
}

@inproceedings{fengtraining,
  title={Training-Free Structured Diffusion Guidance for Compositional Text-to-Image Synthesis},
  author={Feng, Weixi and He, Xuehai and Fu, Tsu-Jui and Jampani, Varun and Akula, Arjun Reddy and Narayana, Pradyumna and Basu, Sugato and Wang, Xin Eric and Wang, William Yang},
  booktitle={The Eleventh International Conference on Learning Representations}
}

@article{lee2023aligning,
  title={Aligning text-to-image models using human feedback},
  author={Lee, Kimin and Liu, Hao and Ryu, Moonkyung and Watkins, Olivia and Du, Yuqing and Boutilier, Craig and Abbeel, Pieter and Ghavamzadeh, Mohammad and Gu, Shixiang Shane},
  journal={arXiv preprint arXiv:2302.12192},
  year={2023}
}

@article{ghosh2023geneval,
  title={Geneval: An object-focused framework for evaluating text-to-image alignment},
  author={Ghosh, Dhruba and Hajishirzi, Hannaneh and Schmidt, Ludwig},
  journal={Advances in Neural Information Processing Systems},
  volume={36},
  pages={52132--52152},
  year={2023}
}

@article{yang2025lora,
  title={Lora-composer: Leveraging low-rank adaptation for multi-concept customization in training-free diffusion models},
  author={Yang, Yang and Wang, Wen and Peng, Liang and Song, Chaotian and Chen, Yao and Li, Hengjia and Yang, Xiaolong and Lu, Qinglin and Cai, Deng and He, Xiaofei and others},
  journal={IEEE Transactions on Image Processing},
  volume={34},
  pages={8145--8158},
  year={2025},
  publisher={IEEE}
}

@inproceedings{lee2026dreamcatcher,
  title={DreamCatcher: Efficient Multi-Concept Customization via Representation Finetuning},
  author={Lee, Jungwon and Lee, Changhun and Park, Eunhyeok},
  booktitle={Proceedings of the IEEE/CVF Winter Conference on Applications of Computer Vision},
  pages={7062--7072},
  year={2026}
}

@inproceedings{
  wang2025msdiffusion,
  title={{MS}-Diffusion: Multi-subject Zero-shot Image Personalization with Layout Guidance},
  author={Xierui Wang and Siming Fu and Qihan Huang and Wanggui He and Hao Jiang},
  booktitle={The Thirteenth International Conference on Learning Representations},
  year={2025},
  url={https://openreview.net/forum?id=PJqP0wyQek}
}

@inproceedings{
jang2024identity,
title={Identity Decoupling for Multi-Subject Personalization of Text-to-Image Models},
        author={Sangwon Jang and Jaehyeong Jo and Kimin Lee and Sung Ju Hwang},
        booktitle={The Thirty-eighth Annual Conference on Neural Information Processing Systems},
        year={2024},
        url={https://openreview.net/forum?id=tEEpVPDaRf}
        }

@article{garibi2025tokenverse,
  title={Tokenverse: Versatile multi-concept personalization in token modulation space},
  author={Garibi, Daniel and Yadin, Shahar and Paiss, Roni and Tov, Omer and Zada, Shiran and Ephrat, Ariel and Michaeli, Tomer and Mosseri, Inbar and Dekel, Tali},
  journal={ACM Transactions On Graphics (TOG)},
  volume={44},
  number={4},
  pages={1--11},
  year={2025},
  publisher={ACM New York, NY, USA}
}

@inproceedings{
zhong2026modadapter,
title={Mod-Adapter: Tuning-Free and Versatile Multi-concept Personalization via Modulation Adapter},
author={Weizhi Zhong and Huan Yang and Zheng Liu and Huiguo He and Zijian He and Xuesong Niu and Di ZHANG and Guanbin Li},
booktitle={The Fourteenth International Conference on Learning Representations},
year={2026},
url={https://openreview.net/forum?id=6wZsaGILlN}
}

\bio{}
\textbf{Trong-Vu Hoang} obtained his B.Sc. degree in computer science from University of Science, Vietnam, in 2024. He currently is a graduate student at University of Science.
\endbio

\bio{}
\textbf{Quang-Binh Nguyen} obtained his B.Sc. degree in computer science from University of Science, Vietnam, in 2024. He is working at Qualcomm AI Research, Vietnam.
\endbio

\bio{}
\textbf{Thanh-Toan Do} is currently a Senior Lecturer at the Department of Computer Science, Monash University, Australia. He obtained Ph.D. in Computer Science from INRIA, Rennes, France in 2012. Before joining Monash, he was a Lecturer at University of Liverpool. Prior to that, he was a Research Fellow at the Singapore University of Technology and Design, Singapore (2013 - 2016) and the University of Adelaide, Australia (2016 - 2018).
\endbio

\bio{}
\textbf{Tam V. Nguyen} is an Associate Professor at Department of Computer Science, University of Dayton. Prior to that, he was a research scientist and principal investigator at ARTIC research centre, Singapore Polytechnic. He was also an adjunct lecturer at National University of Singapore. He received his PhD degree in National University of Singapore in 2013.
\endbio

\bio{}
\textbf{Minh-Triet Tran} is an Associate Professor at Faculty of Information Technology, University of Science, Vietnam. He currently is the Vice Rector of University of Science. He is also the Head of Software Engineering Laboratory, University of Science. He obtained his B.Sc., M.Sc., and Ph.D. degrees in computer science from University of Science, Vietnam, in 2001, 2005, and 2009, respectively
\endbio

\bio{}
\textbf{Trung-Nghia Le} currently is a Senior Researcher and Lecturer at Faculty of Information Technology, University of Science, Vietnam. He received the B.S, M.S, and Ph.D. degrees in computer science from University of Science, Vietnam and Graduate University for Advanced Studies (SOKENDAI), Japan in 2012, 2014, and 2018, respectively. Before joining University of Science, he was postdoctoral researcher and project assistant professor at Kyushu University, University of Tokyo, and National Institute of Informatics, Japan, respectively.
\endbio


\end{document}